\documentclass[10pt]{article}

\AtBeginDocument{%
  }

\usepackage[utf8]{inputenc}
\usepackage[T1]{fontenc}
\usepackage{lmodern}
\usepackage{geometry}
\usepackage{setspace}
\usepackage{microtype}
\usepackage{hyperref}
\usepackage{graphicx}
\usepackage{authblk}
\usepackage{longtable}
\usepackage{booktabs}
\usepackage{array}
\usepackage{enumitem}
\usepackage{url}
\usepackage{ifthen}
\usepackage{xcolor}
\usepackage{times} 
\usepackage{tgbonum}
\usepackage{caption}

\hypersetup{colorlinks=true, linkcolor=black, urlcolor=black, citecolor=black}

\RequirePackage[
  datamodel=acmdatamodel,
  style=acmnumeric,
  sorting=nyt,
  maxbibnames=50,
  ]{biblatex}

\makeatletter
\renewcommand{\fnum@figure}{Figure \thefigure}

\makeatother

\newcolumntype{P}[1]{>{\raggedright\arraybackslash}p{#1}}

\title{AI+HW 2035: Shaping the Next Decade}

\author[1]{Deming Chen}
\author[2]{Jason Cong}
\author[3]{Azalia Mirhoseini}
\author[4]{Christos Kozyrakis}
\author[3]{\\Subhasish Mitra}
\author[5]{Jinjun Xiong}
\author[6]{Cliff Young}
\author[7]{Anima Anandkumar}
\author[8]{\\Michael Littman}
\author[9]{Aron Kirschen}
\author[10]{Sophia Shao}
\author[11]{Serge Leef}
\author[1]{Naresh Shanbhag}
\author[12]{\\Dejan Milojicic}
\author[13]{Michael Schulte}
\author[14]{Gert Cauwenberghs}
\author[15]{Jerry M. Chow}
\author[16,17]{\\Tri Dao}
\author[18]{Kailash Gopalakrishnan}
\author[19]{Richard Ho}
\author[20]{Hoshik Kim}
\author[3,21]{\\Kunle Olukotun}
\author[22]{David Z. Pan}
\author[23]{Mark Ren}
\author[24,25]{Dan Roth}
\author[26]{\\Aarti Singh}
\author[2]{Yizhou Sun}
\author[14]{Yusu Wang}
\author[27]{Yann LeCun}
\author[15]{Ruchir Puri}

\affil[ ]{%
\begin{tabular}{@{}l l l@{}}
$^1$University of Illinois Urbana-Champaign, 
$^2$University of California, Los Angeles\\ 

$^3$Stanford University,
$^4$NVIDIA, 
$^5$State University of New York at Buffalo\\ 

$^6$Google,
$^7$California Institute of Technology, 
$^8$Brown University, 
$^9$SEMRON\\

$^{10}$University of California, Berkeley, 
$^{11}$Synopsys, 
$^{12}$Hewlett Packard Labs\\

$^{13}$Advanced Micro Devices (AMD), 
$^{14}$University of California, San Diego, 
$^{15}$IBM Research\\

$^{16}$Princeton University, 
$^{17}$Together AI, 
$^{18}$EnCharge AI,
$^{19}$OpenAI, 
$^{20}$SK Hynix\\

$^{21}$SambaNova Systems,
$^{22}$The University of Texas at Austin, 
$^{23}$Agentrys,
$^{24}$Oracle AI\\

$^{25}$University of Pennsylvania,
$^{26}$Carnegie Mellon University, 
$^{27}$New York University
\end{tabular}
}

\date{} 

\begin{document}
\maketitle

\vspace{-0.7cm}
\begin{center}
{\small\textbf{Corresponding Authors:} Deming Chen, \href{mailto:dchen@illinois.edu}{dchen@illinois.edu}; Ruchir Puri, \href{mailto:ruchir@us.ibm.com}{ruchir@us.ibm.com}.}
\end{center}

\begin{abstract}
Artificial intelligence (AI) and hardware (HW) are advancing at unprecedented rates, yet their trajectories have become inseparably intertwined. The exponential growth of large AI models and data-intensive applications demands ever more powerful and efficient hardware acceleration, while breakthroughs in specialized computing platforms, ranging from GPUs, FPGAs, and TPUs to emerging NPUs, analog AI chips, photonic systems, and neuromorphic processors, are redefining the limits of intelligent systems. This virtuous cycle is transforming the landscape of computing, but it also exposes a critical gap: despite rapid co-evolution, the global research community lacks a cohesive, long-term vision to strategically coordinate the development of AI and HW. Today’s algorithms are designed around yesterday’s systems, and tomorrow’s chips are optimized for today’s workloads. This fragmentation constrains progress toward holistic, sustainable, and adaptive AI systems capable of learning, reasoning, and operating efficiently across cloud, edge, and physical environments. At the same time, AI’s energy footprint has reached environmentally and economically unsustainable levels. Training a single frontier model can consume energy comparable to hundreds of households, and AI datacenters increasingly rival nations in power demand. The future of AI depends not only on scaling intelligence, but on scaling efficiency, achieving exponential gains in intelligence per joule, defined as meaningful capability, insight, or task performance per unit of energy, rather than unbounded compute consumption. Addressing this grand challenge requires rethinking the entire computing stack. This vision paper lays out a 10-year roadmap for AI+HW co-design and co-development, spanning algorithms, architectures, systems, and sustainability. We articulate key insights that redefine scaling around energy efficiency, system-level integration, and cross-layer optimization. We identify key challenges and opportunities, including the training–inference divide, infrastructure constraints, heterogeneous integration, and equitable access to advanced hardware. We examine important future trends, from memory-centric and 3D-integrated architectures to self-improving systems, decentralized AI agents, and emerging computing paradigms. We candidly assess potential obstacles and pitfalls, including siloed research, resource inequality, and over-reliance on hardware-only gains and propose integrated solutions grounded in algorithmic innovation, hardware advances, and software abstraction.

\textbf{Looking ahead, we define what success means in 10 years: achieving a 1000× improvement in efficiency for AI training and inference}; enabling energy-aware, self-optimizing systems that seamlessly span cloud, edge, and physical AI; democratizing access to advanced AI infrastructure; and embedding human-centric principles into the design of intelligent systems. Finally, we outline concrete action items for academia, industry, government, and the broader community, calling for coordinated national initiatives, shared infrastructure, workforce development, cross-agency collaboration, and sustained public–private partnerships to ensure that AI+HW co-design becomes a unifying, long-term mission.

\end{abstract}



\maketitle

\section*{Executive Summary}

\subsection*{(1) Reinventing Computing and AI Foundations for 1000× Efficiency}
Achieving a 1000× improvement in AI training and inference efficiency requires deep co-innovation between AI models and hardware architectures. The rapid growth of large models has made data movement the dominant bottleneck, outpacing advances in compute, memory, and interconnect technologies. Addressing this challenge calls for a shift toward computation immersed in memory, enabled by dense 3D integration of compute and memory to provide ultra-high bandwidth at low energy cost. Meanwhile, developing low-complexity yet high-quality AI models, including hybrid, Shannon-inspired, neuro-inspired, approximate, and probabilistic models, is critical to reducing computational and memory demands without sacrificing accuracy. Hardware-aware models must further adapt to system constraints through techniques such as redundancy reduction, low-rank and low-precision training, and efficient test-time scaling. Combined with cross-layer optimization and transparent, hardware-agnostic benchmarking frameworks, this tight co-evolution of models, compilers, runtimes, libraries, architectures, and devices can deliver future AI systems that maximize intelligence per joule and usher in a new era of sustainable AI computing.

\subsection*{(2) Revolutionizing Design Productivity and Adaptability}
The pace of AI innovation now far outstrips the speed of hardware and system design. Bridging this gap calls for AI-in-the-loop design workflows that embed learning and reasoning into every stage of development. Open datasets and standardized benchmarks are critical for transparency, reproducibility, and progress in electronic design automation (EDA). Fine-grained task–agent alignment, leveraging specialized large and small language models, will automate and accelerate design subtasks while aiming for intelligence efficiency. Combined with context engineering techniques, these advances will enable AI-native design methodologies that unify technology, architecture, and algorithms into a cohesive, adaptive co-design ecosystem.

\subsection*{(3) Building Reliable and Trustworthy AI Systems}
As AI becomes ubiquitous, reliability and trustworthiness must be understood through fundamental trade-offs among accuracy, robustness, and efficiency, including complexity, energy, and latency. Robustness must span both models and hardware, motivating design methods that explicitly manage these trade-offs and provide guarantees on system behavior. AI hardware paradigms should be evaluated by their position on multi-dimensional trade-off surfaces, with strong methods approaching Pareto-optimality across key metrics. Achieving this requires formal verification, physics-informed constraints, and runtime monitoring. While general-purpose generative AI has transformed many domains, bridging the gap to hardware design demands specialized language models and context-engineered AI systems that understand the semantics of circuits, architectures, and design automation. Benchmarking must also evolve beyond MLPerf to include robustness, explainability, and sustainability.

\subsection*{(4) Physical AI for Scientific Discovery, Robotics and Autonomous Agents}
The next leap in AI innovation lies in coupling data-driven learning with the laws of physics. Physics-informed AI, including approaches based on neural operators and differentiable simulators, offers a principled way to model multi-scale phenomena central to science and engineering, from materials discovery to chip design. At the same time, physical and embodied AI systems, such as robotics and autonomous agents operating in the real world, place stringent demands on energy efficiency, real-time responsiveness, and robustness, making tight integration between learning, control, and hardware essential. Despite their promise, progress in these areas is hindered by the lack of unified benchmarks, datasets, and scalable solvers. Emerging latent world models, such as Joint Embedding Predictive Architectures (JEPA), aim to learn structured latent representations of the physical world. These approaches may provide a foundation for integrating symbolic reasoning, physics-informed priors, and more efficient decision-making mechanisms in future AI systems.

\subsection*{(5) Addressing Core Bottlenecks and Unifying AI+HW Evolution}
A major frontier for future AI lies in developing compact, energy-efficient models that rival frontier models in capability while operating effectively on edge and embedded platforms, including those supporting physical AI. Achieving this will require continued innovation beyond today’s dominant implementations, combining attention mechanisms with complementary architectures, algorithmic sparsity, state-space models, and system-level optimizations to improve efficiency, scalability, and generalization. On the hardware side, the next generation of AI computing platforms will be built on heterogeneous, memory-centric architectures that integrate AI accelerators, programmable fabrics, and quantum processors through scalable, low-latency interconnects. Cross-cutting priorities include AI+HW co-design, energy optimization across the full stack, AI-driven chip and system automation, and large-scale fleet efficiency. In addition, as agentic AI systems become increasingly capable, Human-AI Interaction (HAI) must remain a central focus, ensuring that humans and intelligent agents collaborate seamlessly, communicate intent transparently, and execute complex tasks reliably and safely.

\subsection*{(6) AI and HW in Action: Toward Coordinated Global Impact}
While “AI and HW in Action” may appear industry-focused, academia plays a vital and complementary role in shaping sustainable and globally competitive AI ecosystems. Industry advances rapidly within the prevailing paradigm of large language models (LLMs) and datacenter-scale infrastructure; however, this emphasis can constrain exploration of fundamentally new directions where academia excels. A resilient AI ecosystem relies on academic rigor and critical evaluation to ensure that proposed advances translate effectively into real-world applications. Coordinated AI+HW efforts across the full technology stack are essential to address systemic challenges, including scaling pilot systems to sustained deployment amid regulatory and data sovereignty constraints, managing the escalating cost and energy demands of frontier models, and bridging the gap between open-ended academic inquiry and narrowly scoped industrial objectives. Through aligned policies, shared resources, and sustained collaboration, AI and HW innovation can advance in a sustainable, equitable, and globally impactful manner.

\subsection*{(7) Forging Sustainable Academia–Industry-Government Partnerships}
Achieving the ambitious goals of this vision requires deep collaboration among academia, industry, and government. Expanding government initiatives such as the National Artificial Intelligence Research Resource (NAIRR) will democratize access to compute, data, and models. A persistent challenge remains in bridging academia’s long-term exploratory research with industry’s short-term, product-driven development. Addressing this divide requires shared infrastructure, open-source collaboration, and policy frameworks that align academic creativity with industrial scale and focus, ensuring that innovation remains both foundational and impactful. 

\section{Background and Motivation}

Artificial intelligence has entered an era of unprecedented capability, but also one of profound imbalance. The exponential scaling of AI models, fueled by larger datasets, deeper networks, and vast compute resources, has delivered extraordinary breakthroughs across science, engineering, and daily life. Yet this trajectory is becoming increasingly unsustainable. Each new generation of frontier models demands orders of magnitude more energy and memory bandwidth, with single training runs consuming millions of kilowatt-hours and producing significant carbon emissions. The very systems enabling AI’s ascent are now constrained by the physical, architectural, and economic limits of existing hardware paradigms.

Today’s computing infrastructure remains fundamentally compute-centric, where computation and data storage are disjoint. The resulting “memory wall” creates a severe performance bottleneck, as energy spent on moving data now exceeds that spent on computing it. While GPUs, TPUs \cite{Jouppi_2017}, NPUs, and neuromorphic processors have propelled AI forward, their architectures remain optimized for particular workloads rather than adaptable, evolving computation. Without a unifying AI+HW co-design philosophy, innovation becomes fragmented: AI algorithms assume static backends, and hardware is built for models that quickly become obsolete. This mismatch stifles progress, leading to inefficiencies that compound across the entire system stack.

Meanwhile, software frameworks and algorithmic advances are evolving at a pace that outstrips hardware development cycles by years. This misalignment leaves researchers struggling to adapt rapidly evolving AI paradigms to fixed hardware platforms, while chip designers must anticipate workloads that have yet to emerge. Although the research community has begun to explore AI+HW co-design, as exemplified by \cite{Genc_2021,Hao_2019,Li_2020,Zhang_2022}, such efforts remain far from mainstream within both the AI and hardware communities. The lack of systematic co-design has led to persistent silos, in which data movement, energy optimization, and programmability are treated as separate concerns rather than integrated elements of a coherent system.

The path forward requires rethinking “scaling” itself. Rather than pursuing brute-force computation, the field must embrace energy-aware, self-optimizing, and architecturally adaptive systems. Emerging directions such as memory-centric architectures, dense 3D integration, and compute-in-memory technologies promise to reduce the energy and latency penalties of data movement, as advocated in Illusion \cite{Radway_2021}, N3XT \cite{Srimani_2023}, and Megatrends \cite{IEEE_2024}. Similarly, AI-in-the-loop hardware design, generative EDA tools, and cross-layer optimization frameworks will enable systems that learn, adapt, and co-evolve with the algorithms they support.

In this new paradigm, success is measured not by FLOPs or model size, but by intelligence per joule, trustworthiness, and adaptability across scales. Achieving this vision requires a coordinated, multi-layered roadmap that unites application demands, algorithmic innovation, and enabling hardware technologies into a single, evolving ecosystem. Only through such radical integration can we sustain AI’s growth while ensuring its efficiency, accessibility, and global benefit.

Over the past several years, a vibrant research ecosystem has emerged to support the rapidly growing intersection of AI and HW. New conferences and communities dedicated to AI+HW research, such as MLCAD (Machine Learning for CAD), MLSys (Machine Learning and Systems), and ICLAD (International Conference on LLM-Aided Design), have gained significant traction, bringing together researchers from machine learning, computer architecture, EDA, and systems communities. In addition, many established venues, including flagship conferences such as DAC, MICRO, ISCA, ASPLOS, and ISSCC, increasingly feature dedicated tracks, workshops, and tutorials on AI-driven hardware design and hardware-aware AI algorithms. These developments reflect a rapidly expanding interdisciplinary community and highlight the growing recognition that future breakthroughs will require tight integration between AI algorithms, hardware architectures, and system software. Strengthening and coordinating this research ecosystem will be essential for accelerating innovation and enabling the long-term vision of AI+HW co-evolution.

\section{The Case for Radical Change}

Scaling AI must no longer be defined by more compute, but by better compute. Future AI systems must be energy-aware, self-optimizing, and architecturally adaptive. The shift from compute-centric to memory-centric and data-centric architectures will be essential to overcome the long-standing memory and performance walls that limit today’s systems. To achieve this transformation, innovation must occur across all layers of the computing stack, from materials and devices to algorithms and applications. Equally important, these layers must evolve together through AI+HW co-design and co-evolution, ensuring that efficiency, scalability, and design productivity advance in concert.

\subsection{A Multi-Layered Vision for AI+HW Co-Evolution}
AI’s future depends on a deep, structural rethinking of the relationship between hardware and intelligence. The next decade of progress will not come from isolated breakthroughs but from synergistic innovation across three abstraction layers:
\begin{enumerate}
    \item \textbf{Hardware Technologies} (Hardware Layer)
    \item \textbf{Algorithms and Paradigms} (Algorithm Layer)
    \item \textbf{Applications and Societal Impact} (Application Layer)
\end{enumerate}

\begin{figure}
    \centering
    \includegraphics[width=0.95\textwidth]{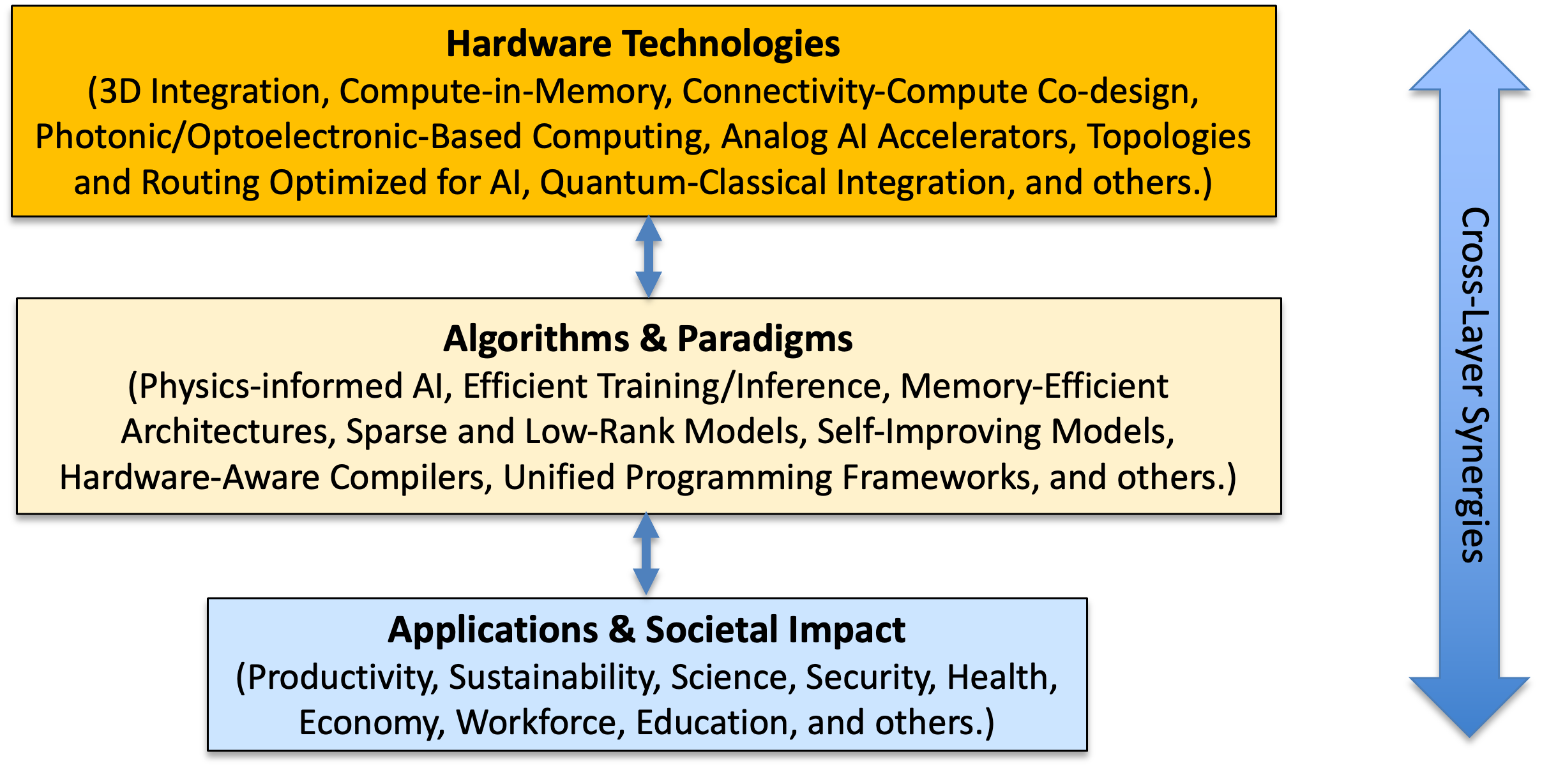}
    \caption{A Multi-Layered Vision for AI+HW Co-Design and Co-Evolution}
    \label{fig:1}
\end{figure}

Hardware Technologies, Algorithms and Paradigms, and Applications and Societal Impact together form a tightly coupled, dynamic feedback loop. At the top, advances in hardware technologies define the performance, energy, and scalability boundaries of AI systems and shape what algorithmic techniques are feasible. Building on these capabilities, the middle layer of algorithms and paradigms translates hardware constraints and opportunities into efficient learning, reasoning, and optimization methods. At the bottom, applications and societal needs drive new requirements on both algorithms and hardware, closing the loop by motivating further innovation across the stack. This strong interdependence calls for continuous cross-layer co-design, in which AI helps design hardware, hardware accelerates AI training and reasoning, and both co-evolve in response to societal priorities such as productivity, sustainability, security \cite{leef2014SECURITY, li2016MLSECURITY}, and reliability \cite{prakash2025DATACENTER, Rossi_2025}. Figure 1 illustrates this multi-layered vision of AI+HW co-evolution. 

\subsection{Key Features of Each Abstraction Layer}
In this subsection, we elaborate on the defining characteristics and design principles of each abstraction layer illustrated in Figure 1. Collectively, these layers capture both the hierarchical structure and the bidirectional interactions that underpin the proposed AI + HW co-design framework. 

\textbf{Hardware Layer: Hardware Technologies.} At the foundation, the coming decade demands radical innovation in hardware design, while working with more efficient AI models, to achieve thousand-fold improvements in AI training and inference efficiency. Key directions include memory-centric architectures that minimize the energy and latency cost of data movement; dense 3D monolithic integration that vertically stacks compute, memory, and interconnect layers; 3D die stacking; and compute-in-memory and analog AI accelerators that perform computation directly where data resides \cite{Wan_2022}. Equally important is connectivity and compute co-design, encompassing photonic and optoelectronic interconnects and computing fabrics that deliver ultra-high bandwidth, low latency, and energy-efficient communication at scale. AI-optimized system topologies and routing strategies are needed to match emerging model structures and dataflows, enabling efficient coordination across heterogeneous components such as CPUs, GPUs, FPGAs, ASICs, and domain-specific accelerators \cite{schulte2015EXASCALE}. In parallel, quantum–classical integration opens new opportunities for hybrid systems that combine classical AI pipelines with quantum processors to address optimization, simulation, and learning tasks beyond classical reach. Together, these advances will redefine the limits of throughput, cost, and power efficiency. Realizing this vision requires AI-driven electronic design automation to become a core part of the hardware workflow, leveraging large and small language models to automate design space exploration, code generation, verification, synthesis, and system-level co-optimization across devices, architectures, and interconnects. Section 3 provides a comprehensive discussion of this abstraction layer, including its key concepts, challenges, emerging trends, potential pitfalls, success milestones over the next decade, and recommended action items for academia, industry, and government.

\textbf{Algorithm Layer: Algorithms and Paradigms.} At this layer, working with the hardware layer, AI must become hardware-aware while hardware becomes AI-adaptive. Today’s decoupling between model innovation and hardware development creates a persistent mismatch: AI evolves on the scale of months, while hardware evolves over years. Closing this gap requires embedding AI directly into the system design loop. AI-in-the-loop design automation will transform how architectures, compilers, and systems are conceived, allowing learning-based methods to optimize memory hierarchies, interconnects, and microarchitectures in near real time. In parallel, hardware-aware training paradigms will improve efficiency through low-precision computation, sparsity, modularity, and memory-efficient execution. Emerging directions such as differentiable simulators, neural architecture search for accelerators, and reinforcement learning based hardware tuning point toward self-evolving computing stacks that continuously improve through feedback. Beyond optimization, new learning paradigms, including physics-informed learning and latent world models such as JEPA \cite{assran2023JEPA, assran2025VJEPA2}, promise AI systems that reason about physical processes rather than merely fitting data, bridging symbolic reasoning with continuous learning. Section 4 examines this abstraction layer in detail, following the same structural framework established in Section 3. 

\textbf{Application Layer: Applications and Societal Impact.} At this layer, AI systems must ultimately serve human and planetary needs while remaining computationally sustainable. As AI permeates productivity tools, health care, materials discovery, education, transportation, economic systems, and national security, computational demand and energy consumption increasingly risk outpacing available infrastructure. By the end of this decade, training a single frontier model could consume energy comparable to that used by entire nations, raising profound environmental, economic, and ethical concerns. Innovations in the hardware and algorithm layers are therefore essential for this layer: advances in hardware technologies enable deployment at scale, while algorithmic innovations, including hardware-aware training, domain-specific models, and physics-informed learning, translate these capabilities into practical, efficient solutions. At the same time, the requirements of this application layer must feed back into the other two layers. Real-world applications demand constraints on energy, latency, robustness, interpretability, and cost, which in turn drive new algorithmic paradigms and impose concrete design targets on hardware systems. Addressing these challenges requires not simply more hardware, but smarter, purpose-driven AI systems. Domain-specific AI, tailored for scientific discovery, engineering design, or physical modeling \cite{li2021NEURALOPERATOR}, can dramatically reduce compute and data requirements by embedding physical laws, structure, and causal priors directly into learning. Many applications will also rely on hybrid edge–cloud architectures, where low-latency reasoning occurs at the edge and large-scale training and adaptation occur in the cloud \cite{chen2025IIDAI}. From an economic and societal standpoint, the metric of success must shift from raw throughput to intelligence per joule. Such a shift would redefine how we evaluate innovation and align technological progress with global sustainability goals. Section 5 provides a focused discussion of this abstraction layer, following the same organizational principles as Sections 3 and 4. 

\subsection{Cross-Layer Co-Design: From Silos to Synergy}

As mentioned before, the future transformations will come not only from advances within layers, but also from cross-layer co-design. Algorithms must adapt to physical constraints; hardware must evolve to serve learning dynamics; and system software must act as the connective tissue ensuring adaptability and reliability.

For example, optimizing end-to-end energy use requires unified abstractions linking model structure to chip layout, runtime scheduling, and even cooling strategies. Similarly, reliability and trustworthiness must be built into the hardware level through formal verification, physics-informed resilience, and secure computation, rather than treated as software afterthoughts.

This vision also redefines design productivity. By leveraging AI models for hardware generation \cite{Dupuis_2025,Sohrabizadeh_2022,Vungarala_2025,Zhang_2020}, verification, and simulation, the cycle from concept to prototype can shrink from years to months or even weeks. Open datasets, modular simulators, and standardized benchmarks, e.g., ITBench \cite{Jha_2025}, IMC-Bench \cite{Shanbhag_2023}, and CVDP Bench \cite{Pinckney_2025}, will further accelerate reproducible progress.

    \begin{center}
    \begin{longtable}[H] {>{\raggedright\arraybackslash}p{2.75cm}>{\raggedright\arraybackslash}p{6.9cm}>{\raggedright\arraybackslash}p{5.7cm}}
    \caption{Fine-Grained Layers with Their Enabling Technologies, Trends, Impacts, and Challenges}
    \label{tab:1}\\
    \hline
    \\[0.05ex]
    \textbf{Layer} &
    \textbf{Key Enabling Technologies / Trends} &
    \textbf{Why It Matters / Challenges} \\
    \\[0.05ex]
    \hline
    \\[0.05ex]
    \mbox{Device} / \mbox{Substrate} / \mbox{Materials} &
    \vspace{-0.3cm} \begin{itemize} [nosep, leftmargin=*]
    \item New materials beyond CMOS (e.g., gallium nitride, carbon nanotubes, 2D materials, silicon-compatible mixed ionic-electronic devices)
    \item Advanced photonic/ optoelectronic in-network computing and integration
    \item Cryogenic / superconducting devices (for quantum / neuromorphic systems)
    \item Non-volatile memory integration beyond MRAM, RRAM, PCM \cite{demasius2021MEMCAPACIOR}
    \end{itemize} &
    Challenges in manufacturing yield, device variability, reliability, cost, and ecosystem maturity. In particular, non-linearity, noise, and drift remain largely unresolved barriers for large-scale crossbar implementations in many emerging non-volatile memory technologies, limiting accuracy, scalability, and long-term stability. \\
    \\[0.05ex]

    \mbox{3D} \mbox{Integration} / \mbox{Heterogeneous} \mbox{Packaging} &
    \vspace{-0.28cm} \begin{itemize} [nosep, leftmargin=*]
    \item 3D monolithic integration, 3D die stacking, chiplets, vertical integration, low-overhead interface protocols
    \item High-bandwidth interposer / interconnect layers
    \item Heterogeneous integration (logic + memory + accelerators in stack)
    \end{itemize} &
    Managing off-chip communication (the “memory wall” bottleneck) is a challenge. Realizing the full potential of 3D designs faces substantial challenges, including thermal management, power delivery, yield and reliability, design complexity, testability, cost / bit, and the need for standardized, low-latency interconnect and packaging ecosystems. \\
    \\[0.05ex]

        \mbox{Analog} / \mbox{Mixed-Signal} / \mbox{In-Memory} \mbox{Compute} &
    \vspace{-0.28cm} \begin{itemize} [nosep, leftmargin=*]
    \item Analog and mixed-signal front-ends to convert physical signals (light, sound, electromagnetic waves, biological signals, etc.) into the digital domain for AI processing.
    \item Compute-in-memory (CIM) or near-memory computing with statistical error compensation methods to enhance accuracy.
    \item Analog AI accelerators (e.g., memristors, resistive crossbars)
    \item Hybrid digital-analog circuits for matrix multiplication
    \end{itemize} &
    These can drastically improve compute efficiency and reduce data movement cost, which dominates energy in large models. However, they bring noise, nonlinearity, precision, calibration issues. Efficient error compensation methods, including statistical ones, need to be developed. \\
    \\[0.05ex]

    \mbox{Photonic} / \mbox{Optical} \mbox{Interconnect} / \mbox{Compute} &
    \vspace{-0.28cm} \begin{itemize} [nosep, leftmargin=*]
    \item Photonic interconnects for chip-to-chip, on-chip
    \item Photonic accelerators (optical matrix multiplication)
    \item Electro-optic co-design, in-network optical computing
    \end{itemize} & 
    Photons are low latency, high bandwidth, and can reduce power in some regimes. Integration with electronics is a major challenge. \\
    \\[0.05ex]

    \mbox{Cooling} \mbox{\&} \mbox{power} \mbox{delivery} &
    \vspace{-0.28cm} \begin{itemize} [nosep, leftmargin=*]
    \item Direct liquid, immersion, and microfluidic cooling for high power density systems.
    \item Power delivery innovations through on-die regulation, backside power delivery, and improved PDN design.
    \item Joint thermal and power optimization with AI-driven runtime management.
    \end{itemize} &
    Power and thermal limits now directly cap AI performance, scalability, and reliability. The main challenges are system complexity, high deployment cost, integration with 3D packaging, and the need for coordinated hardware–software control to manage heat and power safely and efficiently. \\
    \\[0.05ex]

    \mbox{Accelerator} \mbox{Architecture} & 
    \vspace{-0.28cm} \begin{itemize} [nosep, leftmargin=*]
    \item Domain-specific AI accelerators (e.g., tensor cores, NPUs, IPUs)
    \item Sparsity, quantized, low-precision architectures
    \item Reconfigurable architectures (FPGA-like, CGRAs)
    \item Heterogeneous architectures (CPU + GPU + TPU + NPU combinations)
    \end{itemize} & 
    Tailoring architectures to AI workloads yields major gains. However, programmability and flexibility remain key challenges. \\
    \\[0.05ex]

    \mbox{Memory} \mbox{\&} \mbox{Storage} \mbox{Hierarchies} &
    \vspace{-0.28cm} \begin{itemize} [nosep, leftmargin=*]
    \item High Bandwidth Memory (HBM) and its successors
    \item Unified memory models across CPU / accelerator
    \item Emerging memory + flash hybrids (e.g., high bandwidth flash)
    \item Memory compression, caching, tiering strategies
    \item Persistent memory (e.g. NVRAM)
    \end{itemize} & 
    The “memory wall” is a perennial challenge. Keeping highest bandwidth close to compute is critical, while cheaper, slower tiers back up large models. Working with 3D integration to mitigate such a “memory wall” is essential. \\
    \\[0.05ex]

    \mbox{Interconnect} \mbox{\&} \mbox{Networking} &
    \vspace{-0.28cm} \begin{itemize} [nosep, leftmargin=*]
    \item High-throughput, low-latency, duty-cycled, workload-aware  interconnects 
    \item Topologies and routing optimized for AI (all-to-all communications and collective operations)
    \item Interconnect and network optimization for distributed memory coherence / consistency in AI clusters
    \end{itemize} & 
    As models run across many chips or nodes, interconnect latency, bandwidth, and coordination overhead become dominant. \\
    \\[0.05ex]

    \mbox{System} / \mbox{Infrastructure} &
    \vspace{-0.28cm} \begin{itemize} [nosep, leftmargin=*]
    \item Co-design of connectivity and compute
    \item Rack-scale AI systems and modular systems
    \item Disaggregated resources (memory, storage, accelerators)
    \item Power delivery, cooling, energy-efficient design
    \item Edge / on-device AI platforms (tiny AI)
    \end{itemize} & 
    The system-level constraints (power, heat, form factor) often force architectural trade-offs. Reliability could be off by up to 100x if not handled well (e.g., due to silent data corruption). \\
    \\[0.05ex]

    \mbox{Compiler} / \mbox{Runtime} / \mbox{Software} \mbox{Stack} & 
    \vspace{-0.28cm} \begin{itemize} [nosep, leftmargin=*]
    \item Hardware-aware compilers, auto-tuners
    \item Operator libraries (e.g. optimized kernels)
    \item Unified programming frameworks (cross-accelerator)
    \item Co-design of software / hardware (knowing hardware at compile time)
    \item Runtime scheduling across heterogenous units to optimize latency, throughput, or cost
    \end{itemize} & 
    Poor software support causes severe underutilization in heterogeneous, memory-centric systems. Key challenges include performance portability, dynamic scheduling across diverse accelerators, capturing data movement costs, and adapting to power, thermal, and fleet-scale constraints. \\
    \\[0.05ex]

    \mbox{Programming} \mbox{Abstractions} / \mbox{APIs} / \mbox{Models} &
    \vspace{-0.28cm} \begin{itemize} [nosep, leftmargin=*]
    \item Standard APIs \item Model abstraction / intermediate representation strategies, hierarchical representations for planning
    \item Domain-specific languages for AI
    \item Model-to-hardware compilation (i.e. lowering models into hardware instructions)
    \end{itemize} & 
    A major mismatch between current abstractions and heterogeneous, memory-centric systems. Challenges include balancing portability with hardware awareness, exposing structure and locality for optimization, supporting agentic workflows, and enabling model-to-hardware lowering without sacrificing productivity or correctness. \\
    \\[0.05ex]

    \mbox{Algorithms} / \mbox{Models} &
    \vspace{-0.28cm} \begin{itemize} [nosep, leftmargin=*]
    \item Optimal algorithms for navigating the Accuracy-Robustness-Complexity (ARC) trade-offs
    \item Sparse, low-rank, and structured models
    \item Quantization, pruning, and advanced model compression
    \item Efficient training / inference (e.g. federated, on-device, incremental, continual learning)
    \item Memory-efficient architectures
    \item Algorithmic innovations to reduce compute / memory per task
    \item Self improving models
    \item Uncertainty-aware and uncertainty-calibrated models 
    \end{itemize} & 
    Hardware gains alone won’t keep up with model scale. Algorithms must evolve to use compute and memory more efficiently. Challenges include developing hardware-aware, modular, and robust models; enabling small, domain-tuned models to rival large ones; tolerating approximation; and optimizing for memory locality, communication, and heterogeneous execution. \\
    \\[0.05ex]
    \hline
    \end{longtable}
    \end{center}

To expand the vision illustrated in Figure 1 with more concrete and fine-grained detail, Table 1 presents a set of fine-grained abstraction layers together with their enabling technologies, emerging trends, and associated impacts and challenges. It provides a comprehensive and forward-looking view of future AI and hardware systems, spanning from devices and materials to algorithms and models, and highlights why progress at any single layer alone is insufficient to meet the demands of next-generation AI systems. The novelty of the table lies in making explicit the tight coupling and mutual dependencies across layers such as materials, 3D integration, analog and photonic computing, architectures, interconnects, system infrastructure, software stacks, and algorithms, thereby revealing rich cross-layer optimization opportunities that are often obscured by siloed research efforts. By systematically connecting enabling technologies and AI software and hardware to their corresponding impacts and challenges, the table offers a unifying framework to guide academia toward high-impact, interdisciplinary research, help industry prioritize effective co-design strategies for performance, energy efficiency, and scalability, and inform government funding agencies about where sustained, coordinated investments are the most critical. Over the next decade, this layered perspective provides strong strategic value by clarifying how breakthroughs in one layer must be co-developed with innovations in others to overcome fundamental bottlenecks such as data movement, memory walls, power delivery, programmability, and reliability, ultimately enabling sustainable, efficient, and scalable AI systems through holistic cross-layer co-design.

\section{Hardware Technologies for Future AI Systems}

\subsection{Key Insights}

Hardware technologies define the physical limits and opportunities of future AI systems and therefore must be designed in continuous coordination with the algorithmic paradigms (to be discussed in Section 4). As summarized across multiple rows in Table 1 (e.g., 3D Integration / Heterogeneous Packaging; Analog / Mixed-Signal / In-Memory Compute; Photonic / Optical Interconnect / Compute; Cooling and Power Delivery; System Infrastructure), the dominant constraints on AI systems are shifting from raw compute capability to data movement, connectivity, energy efficiency, system-level integration, and cost effectiveness (e.g., \$ / token).
Several fundamental insights emerge:

\begin{itemize}
\item \textbf{System-level constraints have become the primary limiting factors:} Power delivery, cooling, reliability, and data movement now outweigh chip-level considerations, necessitating coordinated co-design across racks and entire compute fleets.
\item \textbf{Data movement has become the primary bottleneck:} The energy cost of moving data across memory hierarchies and interconnects now far exceeds that of arithmetic operations, directly motivating novel algorithmic techniques such as sparsity, locality-aware models, and modular execution as described in Section 4.
\item \textbf{Connectivity is as critical as computation:} Performance scaling increasingly depends on interconnect bandwidth \cite{venkataramani2019SCALINGDATACENTER}, latency, and topology, requiring connectivity–compute co-design rather than treating networking as a secondary concern.
\item \textbf{Integration density reshapes architecture:} Dense 3D integration and heterogeneous packaging collapse traditional boundaries between logic, memory, and interconnect, enabling new algorithmic dataflows that are impossible to achieve on planar systems.
\item \textbf{Hardware must become adaptive:} Fixed-function hardware cannot keep pace with rapidly evolving AI algorithms; instead, hardware must be reconfigurable, programmable, and designed with algorithmic evolution in mind.
\item \textbf{AI must help design hardware:} The scale and complexity of future systems demand AI-driven EDA, creating a closed feedback loop in which AI systems design the very hardware that accelerates future AI models.
\end{itemize}

These insights underscore that hardware innovation and algorithmic innovation are inseparable and must be co-designed as a unified system.

\subsection{Key Challenges and Opportunities}

The hardware layer faces several intertwined challenges that simultaneously create unprecedented opportunities for innovation when viewed through a cross-layer lens.

\

\noindent
\textbf{Key Challenges}

\begin{itemize}
\item Memory and Data Movement Wall (Table 1: Memory Hierarchy, Interconnect)

\begin{itemize}
\item Off-chip memory access dominates energy and latency in both training and inference.
\item Conventional cache-based hierarchies poorly match AI access patterns, especially for attention and large embedding tables.
\item These constraints directly motivate algorithmic compression, locality-aware training, and model partitioning strategies.
\end{itemize}

\item Connectivity Scaling Limits (Table 1: Photonic / Optical Interconnect)

\begin{itemize}
\item Electrical interconnects struggle to scale in bandwidth density and energy efficiency.
\item Network startup latency and protocol overhead limit effective utilization for large distributed models.
\item Mismatch between network topology and AI dataflow reduces achievable performance.
\end{itemize}

\item Thermal and Power Delivery Constraints (Table 1: Cooling and Power Delivery)

\begin{itemize}
\item Dense 3D stacks exacerbate heat dissipation challenges.
\item Power and thermal limits now directly cap AI performance, scalability, and reliability.
\item Thermal variability introduces reliability and lifetime concerns.
\end{itemize}

\item Design Complexity Explosion (Table 1: System Infrastructure, EDA)

\begin{itemize}
\item Hardware simulators and evaluation tools increasingly lag behind real-world system complexity, creating a validation gap between research prototypes and deployable systems.
\item The design space now spans devices, materials, packaging, architectures, interconnects, and software. Manual design flows cannot explore this space effectively, leading to suboptimal systems.
\end{itemize}
\end{itemize}

While the hardware layer faces fundamental scaling challenges, it also presents some of the most powerful leverage points for achieving transformative gains when co-designed with algorithms and systems in Section 4.

\

\noindent
\textbf{Key Opportunities}

\begin{itemize}
\item \textbf{Toward Cross-Layer, System-Centric AI Hardware Design.} Future AI hardware must be designed through a cross-layer, system-centric lens that tightly couples algorithms, compilers, and physical platforms. This requires new hardware abstractions that expose data movement, memory locality, and energy cost as first-class primitives, enabling learning algorithms to reason directly about physical constraints rather than relying on abstract FLOPs. Hardware research should also co-evolve with emerging AI paradigms, such as modular, agentic, and physics-informed models, by supporting reconfigurable execution substrates, elastic memory hierarchies, and programmable interconnects that adapt as models change. Correspondingly, evaluation methodologies must move beyond component-level metrics toward end-to-end system measures, including intelligence per joule, sustained utilization under real workloads, and robustness to thermal and power variability, to accurately capture the true effectiveness of AI+HW co-design.
\item \textbf{Memory-Centric and In-Memory Computing as an Algorithmic Enabler.} Analog, mixed-signal, and digital compute-in-memory architectures \cite{Kang_2014,Wan_2022,Zhang_2024} offer a path to fundamentally reducing data movement by colocating computation with storage which today provide state-of-the-art energy efficiency and compute density \cite{Shanbhag_2023} although scaling to larger models remains a significant challenge. Beyond raw efficiency, these architectures enable new algorithmic abstractions. For example, approximate computation~\cite{Kang_2020} and noisy arithmetic can be embraced by algorithms that are inherently robust, probabilistic, or self-correcting, as discussed in Section 4. This opens up opportunities for learning paradigms~\cite{Shanbhag_2019} that trade exactness for orders-of-magnitude efficiency gains, particularly in training and large-scale inference.
\item \textbf{3D Integration and Heterogeneous Packaging for New Dataflows.} Dense 3D integration and advanced packaging collapse the physical distance between logic, memory, and interconnect. This enables fine-grained, bandwidth-rich communication patterns that can be exploited by hierarchical, modular, and locality-aware models. From a co-design perspective, algorithm designers can assume new forms of spatial and temporal locality, while hardware designers can tune vertical integration strategies based on model structure, leading to fundamentally new compute–memory dataflows.
\item \textbf{Photonic and Optoelectronic Connectivity as a Scaling Breakthrough.} Photonic and optoelectronic interconnects provide near distance-independent bandwidth and latency, enabling scale-out and scale-up AI systems that are not bottlenecked by electrical signaling limits. This creates opportunities for algorithmic paradigms that assume abundant, low-latency global communication, such as large-scale model parallelism, distributed attention, and collective reasoning across agents. Over time, photonic computing elements may also enable new primitives for linear algebra and signal processing.
\item \textbf{Connectivity–Compute–Topology Co-Design.} AI workloads exhibit highly structured communication patterns that are poorly matched to traditional network topologies. Co-designing system topology, routing, and compute placement around AI dataflows enables higher utilization and lower energy consumption. This opportunity directly aligns with algorithmic research on structured sparsity, pipeline parallelism, and graph-based execution models \cite{luo2024DEHNN} in Section 4.
\item \textbf{AI-Driven Design Automation as a Force Multiplier.} The sheer complexity of future hardware systems makes AI-driven EDA \cite{Fayyazi_2024,Fu_2023,Lin_2019, Mirhoseini_2021,Mishty_2024,Zhang_2025}, not just beneficial, but essential. Learning-based design tools can explore massive design spaces, optimize trade-offs across layers, and rapidly specialize hardware for emerging algorithms. In turn, these tools rely on advances in learning, optimization, and representation of AI models, creating a virtuous cycle where AI improves hardware that accelerates future AI.
\item \textbf{Thermal Scaffolding for Ultra-dense 3D ICs.} Emerging thermal dielectric materials and thermal scaffolding structures will become increasingly important for ultra-dense 3D integrated circuits, enabling improved heat spreading, reduced thermal resistance, and enhanced reliability in vertically stacked AI systems \cite{rich2023thermal}. Such material-level innovations can fundamentally reshape the thermal envelope of next-generation AI accelerators.
\end{itemize}

\subsection{Key Questions and Answers}

\

\noindent\textbf{Q1: Is hardware innovation still the dominant driver of AI progress, or have algorithms overtaken it?}
\begin{itemize}
\item \textbf{Answer:} Neither alone is sufficient. Algorithmic breakthroughs increasingly depend on hardware capabilities, while hardware gains only translate into impact when matched by algorithmic adaptation. Sustained progress requires continuous co-evolution, where hardware enables new algorithmic paradigms and algorithms actively shape hardware design targets.
\end{itemize}

\noindent\textbf{Q2: Can specialization and generality coexist in future AI hardware?}
\begin{itemize}
\item \textbf{Answer:} Yes, but only through hierarchical and modular design. Specialized accelerators, chiplets \cite{Mishty_2024,Shao_2019}, and analog or photonic components must be composed into flexible systems via programmable interfaces and compiler support. Algorithmic modularity and composability, as discussed in Section 4, are essential to making specialization sustainable rather than brittle.
\end{itemize}

\noindent\textbf{Q3: How much approximation and heterogeneity can AI systems tolerate?}
\begin{itemize}
\item \textbf{Answer:} More than traditional computing models assume. Many AI workloads are inherently statistical and can tolerate noise, reduced precision, and approximate computation. Algorithmic techniques such as robustness-aware training, uncertainty modeling, and adaptive precision enable systems to leverage the statistical nature of AI models and heterogeneous hardware components while preserving accuracy and reliability.
\end{itemize}

\noindent\textbf{Q4: Can hardware design cycles realistically match the pace of AI innovation?}
\begin{itemize}
\item \textbf{Answer:} Not with traditional workflows. However, AI-in-the-loop hardware design, generative EDA, and reusable chiplet ecosystems can dramatically shorten design cycles. This approach mirrors trends in Section 4, where learning systems continuously adapt rather than being statically defined.
\end{itemize}

\noindent\textbf{Q5: How should success be measured at the hardware layer?}
\begin{itemize}
\item \textbf{Answer:} Traditional metrics such as peak FLOPs are insufficient. Success must be measured in terms of system-level outcomes such as intelligence per joule, end-to-end latency, scalability, and adaptability to evolving algorithms and applications. These metrics directly align with the application-level goals discussed in the bottom layer and reinforce the need for cross-layer optimization.
\end{itemize}

\noindent\textbf{Q6: How do societal and application needs influence hardware priorities?}
\begin{itemize}
\item \textbf{Answer:} Requirements such as energy efficiency, robustness, real-time responsiveness, and deployability at the edge must feed back into hardware design. These constraints shape both algorithmic choices and hardware architectures, ensuring that innovation remains aligned with societal impact rather than purely technical benchmarks \cite{chen2021HUMANEVAL,hendrycks2021MMLU}.
\end{itemize}

\subsection{Important Future Trends}

In order to capture the future trends of AI model scaling and efficiency, we study the joint evolution of time, accuracy, and model size using representative data from 2020–2025, with trends extrapolated toward 2030. We present both two-dimensional and three-dimensional projections to reveal complementary insights. Figure 2 shows the two-dimensional plots, which expose clear pairwise relationships among model size, accuracy, and time. Although these visualizations are two-dimensional, additional information from the third dimension is implicitly encoded—for example, in the Model Accuracy versus Time plot, model size is represented by the circle size. From these plots, several consistent trends emerge. First, accuracy improves steadily over time even as model sizes vary, indicating efficiency gains driven by algorithmic and architectural advances beyond brute-force scaling. Second, within a fixed time window, higher accuracy generally correlates with larger model size, reflecting prevailing scaling laws of contemporary foundation models. Third, for a fixed model size range, accuracy increases over time, highlighting progress in training methods, data curation, and model design. While informative, these trends become more complete and intuitive when their correlations are examined directly in a three-dimensional space, as we show in Figure 3. 

\begin{figure}
    \centering
    \includegraphics[width=0.95\linewidth]{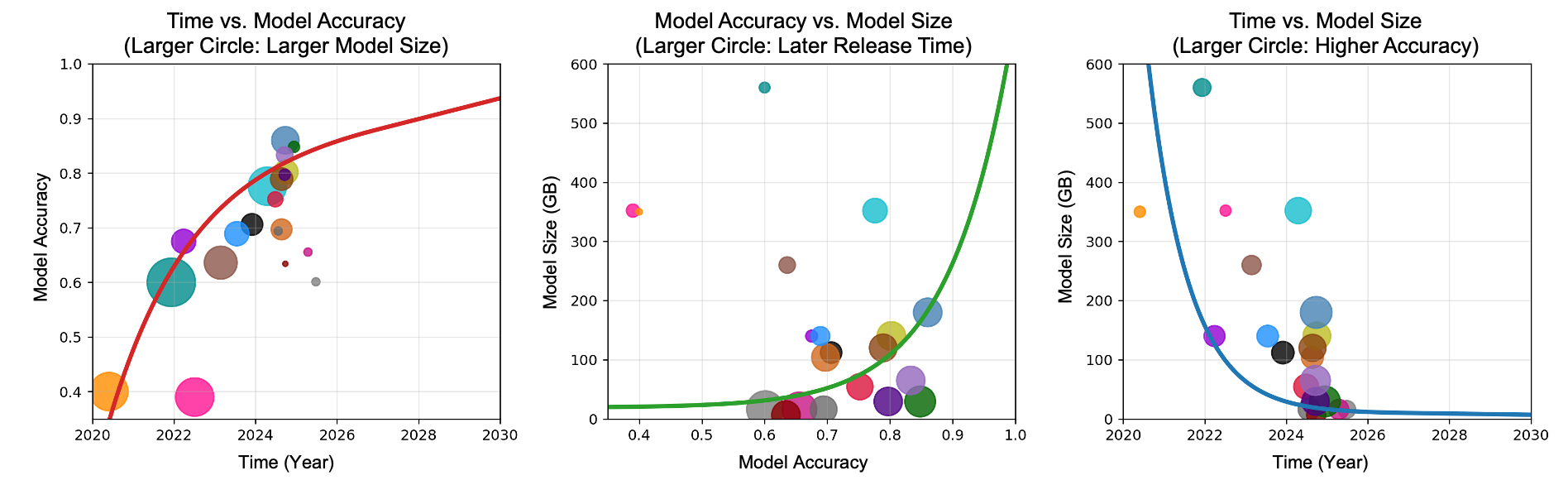}
    \caption{Predicted Two-Dimensional Trends across Model Size, Accuracy, and Time}
    \label{fig:2}
\end{figure}

\begin{figure}
    \centering
    \includegraphics[width=0.7\linewidth]{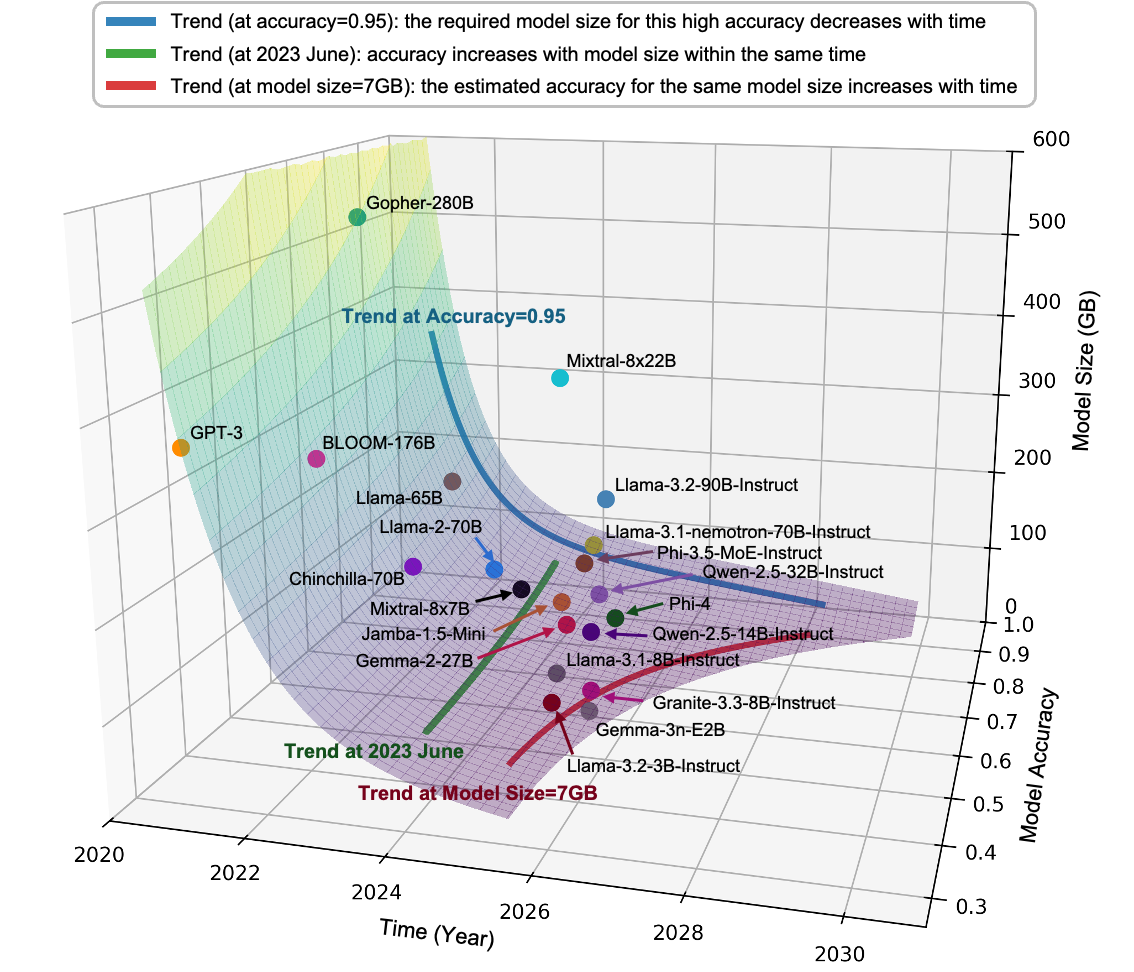}
    \caption{Predicted Three-Dimensional Trends across Model Size, Accuracy, and Time}
    \label{fig:3}
\end{figure}

Figure 3 extends this analysis by integrating time, model size, and accuracy into a single three-dimensional geometric representation, capturing their joint correlations using the same underlying data points. By explicitly placing each model in a shared 3D space, this view enables direct comparison of progress across generations, clearer identification of outliers, and a more holistic assessment of efficiency gains. Importantly, the 3D representation allows one dimension to be fixed, such as time, accuracy, or model size, while observing how the other two dimensions co-evolve, as illustrated by the three curves (blue, green, red) drawn in the 3D plane. This enables additional insights over the 2D projections alone, such as how the model size required to reach a target accuracy shrinks over time, or how accuracy evolves for a fixed resource budget. Beyond individual trends, the 3D view provides a structured and comprehensive picture of the co-evolution of models, performance, and efficiency.

These trends point to a fundamental shift in how progress in AI systems is defined and pursued. Historically, advances were often driven by optimizing a single dominant dimension—most notably model size—under the assumption that larger models would inevitably deliver higher accuracy. While this scaling-driven phase produced remarkable gains, it is now approaching saturation, prompting a natural pivot toward optimizing other critical dimensions such as efficiency, power, latency, cost, and deployability. The long-term trajectory of the field is not focusing on trade-offs among these dimensions, but by convergence toward solutions that would improve the design quality across all dimensions, i.e.,  smaller, more specialized models achieving higher accuracy and efficiency through the co-design of algorithms, hardware, and systems.

In this emerging paradigm, intelligence efficiency would become a central metric, and the narrative will shift from individual models to full systems and agentic ecosystems, where the right model is dynamically matched to the right task. Large models will remain indispensable for complex reasoning \cite{Brown_2020,Feng_2025,Wei_2022} and knowledge synthesis, while smaller, domain-tuned models will dominate focused and resource-constrained workloads, especially as AI increasingly interacts with the physical world. Looking toward 2035, physical AI is expected to account for the vast majority of real-world inference, enabled primarily by highly efficient small models. This outlook reinforces the imperative to design AI systems holistically, with intelligence per joule, per dollar, and per second treated as first-class objectives. Informed by these trends, we highlight the following key directions for enabling technologies.

\subsubsection{Near-Term Trends in Enabling Technologies (2–5 years):}

\

\

\noindent\textbf{Domain-specific AI accelerators (e.g., tensor cores, NPUs) with native support for quantization and sparsity}, enabling substantial gains in performance-per-watt by aligning hardware execution with the structure, precision, and sparsity of modern AI workloads.

\

\noindent\textbf{Heterogeneous compute nodes combining CPUs, GPUs, and NPUs}, allowing different components of AI pipelines—control logic, dense computation, sparse execution, and I/O—to be mapped to the most suitable hardware, improving utilization and reducing system-level inefficiencies.

\

\noindent\textbf{High-bandwidth memory (HBM) integration with wider interfaces and tighter coupling to compute}, addressing the growing dominance of memory bandwidth and data movement as primary bottlenecks in both training and inference.

\

\noindent\textbf{3D packaging and chiplet-based architectures}, enabling scalable composition of compute, memory, and specialized accelerators while improving yield, flexibility, and time-to-market compared to monolithic designs.

\

\noindent\textbf{Hardware-aware compilers, auto-tuners, and optimized operator libraries}, which are essential for translating architectural advances into real, sustained efficiency gains and for closing the gap between peak capability and achieved utilization.

\

\noindent\textbf{Standardization across APIs, intermediate representations, model formats, operator and kernel libraries, and security, privacy, and provenance specifications}, reducing ecosystem fragmentation and enabling portability, interoperability, and faster adoption of heterogeneous AI systems.

\

\noindent\textbf{Edge and on-device AI capabilities centered on small, efficient models}, supporting low-latency, privacy-preserving, and energy-efficient inference, and laying the foundation for large-scale deployment of AI systems interacting with the physical world.

\

\noindent\textbf{Hybrid and unified memory hierarchies}, combining fast DRAM with slower NVRAM or flash and enabling unified memory across CPUs, GPUs, and accelerators to better match the tiered access patterns and persistent state requirements of emerging AI workloads.

\subsubsection{Medium/Long-Term Trends in Enabling Technologies (6–10 years):}

\

\

\noindent\textbf{Quantum-accelerated AI through hybrid quantum–classical systems}, where quantum processors complement classical AI pipelines for optimization, sampling, and simulation tasks that are difficult to scale using conventional approaches.

\

\noindent\textbf{Photonic and optical interconnects within and between chips}, offering orders-of-magnitude improvements in bandwidth density and energy efficiency, and enabling large-scale AI systems to scale beyond the limits of electrical interconnects.

\

\noindent\textbf{Photonic accelerators and analog–optical hybrid computing}, providing new energy-efficient primitives for linear algebra and signal processing in specialized AI workloads.

\

\noindent\textbf{Broader adoption of compute-in-memory and analog computing}, fundamentally reducing data movement by colocating computation with storage, while relying on algorithmic robustness and error-tolerant learning to manage noise and variability.

\

\noindent\textbf{Dense 3D heterogeneous integration of compute, memory, and logic}, collapsing traditional architectural boundaries and enabling new dataflows, tighter coupling, and higher system-level efficiency—capabilities that will be critical for supporting many emerging applications and workloads in future physical AI systems.

\

\noindent\textbf{New materials and transistor technologies beyond conventional CMOS}, if manufacturable at scale, to overcome fundamental limits in power, speed, and integration density.

\

\noindent\textbf{Ultra-scalable distributed AI systems with adaptive coherence and orchestration}, capable of dynamically managing computation, memory, and communication across cloud, edge, and device layers, supporting coordinated operation of large numbers of intelligent agents.

\subsection{Potential Obstacles, Pitfalls, and Solutions}

\noindent\textbf{Obstacles and Pitfalls}
\begin{itemize}
\item Noise, drift, and calibration challenges in analog and photonic systems.
\item Yield and reliability concerns in dense 3D integration.
\item Fragmented software ecosystems that limit portability.
\item Over-specialization that reduces generality and reuse.
\item Silent Data Corruption (SDC) in large-scale AI systems fabricated at advanced technology nodes, including test escapes and fleet-level
latent errors, poses a growing threat to reliable computing, especially under aggressive voltage
scaling and heterogeneous integration. 
\end{itemize}

\noindent\textbf{Potential Solutions}
\begin{itemize}
\item Algorithmic robustness techniques such as noise-aware training and error compensation.
\item Adaptive calibration using embedded learning agents.
\item Modular hardware and software interfaces aligned with algorithmic abstractions in Section 4.
\item Community-driven standards and open benchmarks \cite{bai2025LONGBENCHV2, glazer2025FRONTIERMATH, zuo2025PLANETARIUM}.
\item Cross-layer reliability monitoring, error detection and correction mechanisms, fleet-scale telemetry analysis, and hardware–software co-design approaches that explicitly model and mitigate SDC risks across production deployments \cite{deutsch2026pindrop,mitra2024sdc}.
\end{itemize}

\subsection{What Does Success Mean in 10 Years}

In ten years, hardware success includes seamless interoperability across heterogeneous components, where new accelerators can be integrated without redesigning the entire software stack, and reliability scales predictably with system size. Data movement is minimized by design, connectivity scales transparently, and hardware adapts as algorithms evolve. Analog, digital, photonic, and quantum components coexist within unified systems. Hardware and algorithms continuously co-optimize through AI-driven design automation. Most importantly, these systems deliver dramatically higher intelligence per joule, aligning technological progress with societal and environmental sustainability. 

Success also entails hardware platforms that can be rapidly re-specialized through software and compilation, or structurally reconfigured, enabling new AI models, agents, and physical-world workloads to be deployed without redesigning silicon, thereby closing the longstanding gap between hardware lifecycles and the pace of AI innovation.

\subsection{Suggested Action Items for Academia, Industry, Government, and the Community}

\noindent\textbf{Academia}
\begin{itemize}
\item Lead interdisciplinary research spanning materials, devices, architectures, algorithms, and EDA.
\item Develop open testbeds and benchmarks that reflect cross-layer interactions.
\item Train students to be fluent across hardware and AI boundaries.
\end{itemize}

\noindent\textbf{Industry}
\begin{itemize}
\item Invest in co-design across hardware and algorithms rather than siloed optimization.
\item Share pre-competitive infrastructure and standards.
\item Deploy AI-driven design workflows at scale.
\end{itemize}

\noindent\textbf{Government}
\begin{itemize}
\item Fund long-horizon research in 3D integration, photonics, analog AI, and quantum–classical systems.
\item Support national shared infrastructure and open platforms.
\item Incentivize cross-sector collaboration.
\end{itemize}

\noindent\textbf{Community}
\begin{itemize}
\item Shift evaluation metrics toward system-level efficiency and societal impact.
\item Encourage reproducibility, openness, and interoperability.
\item Foster a culture of holistic, cross-layer innovation.
\end{itemize}

\section{Algorithms and Paradigms for Scalable AI+HW}

\subsection{Key Insights}

This section identifies core bottlenecks to scalable AI across algorithms, architecture, and infrastructure; discusses the need and opportunity of developing smaller efficient models that can rival large ones for edge applications, including physical AI; examines the limits of the current attention-based approaches for LLMs and explores new AI models; and proposes research for future, efficient hardware architectures for AI acceleration with a focus on heterogeneous, coarse-grain reconfigurable, memory-centric compute stack spanning CPUs, GPUs, programmable fabrics, and even quantum processors with scalable interconnects. Cross-cutting themes include AI model and hardware co-design, energy efficiency optimization, AI-driven chip design automation, compute–memory integration, and gigawatt-scale fleet optimization. Physics-informed learning, neural operators, and hybrid symbolic–physical reasoning are critical not only for scientific applications but also for improving efficiency, robustness, and interpretability of AI systems. Agentic AI systems increasingly act as orchestrators, selecting models, kernels, hardware resources, and execution strategies dynamically under real-world constraints.

Algorithmic innovation has historically delivered step-function efficiency gains that rival or exceed those achieved through hardware scaling alone. Past transitions—such as the evolution from recurrent architectures to attention-based and state-space models—demonstrate that fundamental changes in model structure, training dynamics, and representation can unlock scalability and efficiency that were previously unattainable. Looking forward, similar breakthroughs are expected from advances in modular architectures, long-term memory systems, sparsity-aware learning, causal and physics-informed representations, and agentic decomposition of tasks. These algorithmic shifts can dramatically reduce computation, memory traffic, and communication requirements, thereby reshaping hardware design targets rather than merely adapting to them. Consequently, achieving 1000× efficiency gains for AI training and inference will require sustained investment in algorithmic research that redefines what computation is necessary, not just how efficiently existing computations are executed.

Training and inference impose fundamentally different system requirements and must be treated as distinct co-design targets. Training workloads prioritize throughput, statistical efficiency, peak accuracy, and amortized energy cost over long time horizons, whereas inference—particularly for physical AI systems such as robotics, autonomous vehicles, and industrial control—demands millisecond-scale latency, deterministic response, and extreme energy efficiency under tight power budgets. For embodied systems operating continuously in the physical world, energy efficiency directly translates into operational lifetime (e.g., hours of autonomy per charge), safety margins, and thermal reliability. Existing deployed systems, such as autonomous driving platforms, already demonstrate the feasibility of large-scale real-world inference under strict latency and power constraints, offering valuable lessons for AI+HW co-design. These systems reveal that inference efficiency is dominated not only by arithmetic cost but by memory access, sensor fusion, control-loop integration, and worst-case execution guarantees. Looking forward, achieving orders-of-magnitude efficiency improvements for physical AI will require hardware–software stacks optimized explicitly for real-time inference, including predictable memory hierarchies, locality-first execution, mixed-criticality scheduling, and domain-specialized models that balance accuracy, robustness, latency, and energy efficiency.

These have to be achieved with close, cross-disciplinary collaboration, with the goal to narrow and eventually eliminate innovation-speed mismatch between fast-moving models/algorithm development versus slower-pace hardware roadmaps. Human-AI Interaction (HAI) remains a high priority, especially in the agentic era, where human and agents need to collaborate seamlessly for people to express intent and have machines reliably execute complex tasks.

\subsection{Key Challenges and Opportunities}

To enable scalable AI+HW innovation, one needs to address the following challenges:

\begin{itemize} [leftmargin=*]
\item \textbf{Siloed hardware development and model design} (Table 1: Algorithms, Models, Programming Abstraction, System Infrastructure):
\item [] AI hardware development has traditionally been siloed, with algorithms, compilers, and physical platforms optimized largely in isolation and evaluated using narrow, component-level metrics such as peak FLOPs or bandwidth. Future AI hardware must be designed through a cross-layer, system-centric lens that tightly couples algorithms, compilers, and physical platforms. This requires new hardware abstractions that expose data movement, memory locality, and energy cost as first-class primitives, enabling learning algorithms to reason directly about physical constraints. Models based on cross-layer learning will dynamically adapt execution strategies, including precision, sparsity, partitioning, and placement, based on real-time hardware telemetry such as congestion, thermal conditions, and energy availability.
\end{itemize}

\begin{itemize} [leftmargin=*] 
\item \textbf{Algorithmic brute force \& retrieval dominance} (Table 1: Algorithms, Models, Memory Hierarchy, Programming Abstraction): 
\item [] Current models rely on attention, vector similarity, and retrieval as the dominant primitives, leading to the inefficiency associated with increasing parameter counts and context lengths. Human-like abstraction and alternative model architectures, including ensembles of small models, can lead to significant efficiency improvements. New learning algorithms are required to exploit deep memory hierarchies, tiered storage, and persistent memory, shifting optimization targets from FLOPs to memory traffic and data locality. 
\end{itemize}

\begin{itemize} [leftmargin=*] 
\item \textbf{Energy, memory, and interconnect walls} (Table 1: Memory Hierarchy, Interconnect, 3D Integration, Heterogeneous Packaging):
\item [] Energy is the limiting factor at the hardware level, with the interconnects within and across chips being the main source of energy overheads. This leads to memory capacity and bandwidth becoming major performance bottlenecks. Near/in-memory compute, 2.5D/3D heterogeneous integration, and optical interconnects are promising directions to overcome these challenges. Memory-efficient architectures, including those proposed in Mamba \cite{Gu_Dao_2023} and HMT \cite{He_2025}, represent promising directions that should be explored further. 
\end{itemize}

\begin{itemize} [leftmargin=*] 
\item \textbf{Low utilization \& co-design gap} (Table 1: Accelerator Architecture, System Infrastructure, Compiler, Runtime, Software Stack): 
\item [] In real deployments, systems often operate at only ~5–20\% utilization. New accelerator chips frequently arrive before software stacks are fully tuned for the previous generation, highlighting the need for automated, cross-layer design-space exploration and optimization. This co-design approach can bridge the current gap between the pace of AI algorithms and the pace of hardware design. There is a growing opportunity for self-improving systems in which models generate optimized kernels, guide compilation, and refine hardware utilization over time. Hardware-efficient architectures, such as FlashAttention \cite{Dao_2022}, PagedAttention \cite{Kwon_2023}, and RadixAttention \cite{zheng2024SGLANG} have shown strong promise and warrant continued investigation and broader adoption.
\end{itemize}

\begin{itemize} [leftmargin=*] 
\item \textbf{Gigawatt-scale operations} (Table 1: Interconnect \& Networking, System Infrastructure, Cooling \& Power Delivery): 
\item [] Optimization must target the entire GW-scale fleet (scheduling, placement, power / thermal / cooling, grid constraints), not just per node. The goal is to optimize performance per watt and accuracy per watt across the whole deployment. Interconnect-aware and topology-aware models must co-evolve with workload-aware networking fabrics to reduce synchronization and communication overheads at scale.
\end{itemize}

\begin{itemize} [leftmargin=*] 
\item \textbf{Edge constraints} (Table 1: Memory Hierarchy, System Infrastructure, Algorithms, Models): 
\item [] On-device AI for robotics and mobile systems presents both an emerging challenge and a significant opportunity for new hardware innovations (e.g., 3D integration or in-memory computing) and for tightly co-designing application-specific models with specialized hardware.
\end{itemize}

\subsection{Key Questions and Answers}

Below we summarize several central questions related to AI+HW co-innovation at the algorithms and platforms level and provide initial answers. 

\

\noindent\textbf{Q1. What is the bottleneck for scalable AI + HW?}

\noindent\textbf{Bottlenecks:} energy constraints, memory (capacity / bandwidth / locality) walls, interconnect fabric limitations, under-utilized infrastructure, and lack of abstractions. \textbf{Remedies:} fine-grained compute-in / near-memory integration, scalable 3D memory, better design-space exploration for both model and hardware development, earlier co-design between AI research, hardware design, compiler development, and fleet-level (GW-scale) optimization.

\

\noindent\textbf{Q2. Can 10–100× smaller models be equally capable?}

\noindent It is possible with a focus on selected application domains. Possible paths include pruning and quantization; domain-specific distillation with clear legal and IP frameworks; novel architectures such as those incorporating long-term memory; radically more efficient heterogeneous hardware; and hybrid deployment strategies. \textbf{An Ecosystem View:} a government sponsored, community-based effort to train large “teacher” models with explicit permission for distillation; deploy small models where efficiency, latency, or privacy are critical, and large models where quality is paramount; and enable multi-agent systems in which local models collaborate and selectively invoke larger models as needed.

\

\noindent\textbf{Q3. Is attention all we need?}

\noindent No. Attention is central for LLMs (Large Language Models) but not universal; convolutions, SSMs (State Space Models), diffusion models also matter. It is important to use objective, task-relevant metrics and avoid conflating correlation with causation. Today’s LLM inference is largely memory bound due to the massive number of parameters \cite{grattafiori2024LLAMA3, guo2025DEEPSEEKR1, yang2025QWEN3} and the growing dominance of KV cache traffic as context length increases \cite{liu2025LMCACHE}; sliding-window/sparse variants, cache sharing, and use of long-term memories help, but further improvement in arithmetic intensity remains to be critical. Hardware should emphasize reconfigurable, lower-level primitives and consider boosting better memory utilization, not just FLOPs.

\

\noindent\textbf{Q4. What is the ideal hardware architecture?}

\noindent \textbf{Heterogeneous, massively parallel, memory centric systems:} energy-efficient cores tightly coupled with 3D stacked and scalable memory; efficient support for dense local and sparse global connectivity following a small-world network model; optical links for high bandwidth global communication; reconfigurable fabrics for flexibility; and targeted use of quantum computing \cite{Havlicek2019QUANTUMSVM}. These systems must also address deployment complexity, software stack integration, and fleet level power and capital expenditure trade-offs.

\

\noindent\textbf{Q5. What are the top research priorities?}

\noindent Human–AI Interaction (HAI) to bridge human intent and machine execution through clearer abstractions, well-defined human-in-the-loop roles, and effective human–agent collaboration; cross-layer exploration and co-design of AI algorithms, systems, chips, and design workflows using AI-assisted techniques; AI-enabled quantum computing, including quantum error-correction decoding, compilation, and embedded AI operating under cryogenic and control-power constraints; AI-driven chip and system design automation with agentic orchestration, where agents dynamically select models, resources, and contextual information; and self-improving systems in which models generate optimized kernels, continuously refine their own infrastructure, and adapt to domain-specific, often live, data streams.

\subsection{Important Future Trends}

\begin{itemize}
\item \textbf{Converged heterogeneous stacks:} Future AI infrastructure will feature converged heterogeneous stacks \cite{yoon2025LLMSERVING} integrating classical, AI-specific dense computation, reconfigurable fabrics, and quantum computers, with optical global links atop dense local 3D compute-memory.
\item \textbf{Compute–memory convergence} to overcome energy / latency limits. Techniques such as compute-in / near-memory \cite{Wan_2022}, 3D stacking \cite{liu20253DSTACK}, and memory-centric dataflow \cite{Lacouture_2025,Ye_2025} will become mainstream. These changes will require new programming models, thermal-aware designs, and hybrid analog-digital components optimized for locality and efficiency.
\item \textbf{Small–big model symbiosis:} Large models will serve as sources for distillation and reasoning scaffolds, while compact SLMs (Small Language Models) run efficiently on edge and embedded devices. Domain-tuned SLMs will be distilled from open frontier teachers, orchestrated by multi-agent frameworks.
\item \textbf{Mechanistic understanding driving specialization:} As interpretability research reveals how models internally represent computation, this knowledge will translate into new, specialized data structures and domain-optimized kernels. Model insights will guide compression, caching, and sparsity strategies, and will increasingly be compiled directly into hardware instructions, enabling automated generation and formal verification of kernels for both performance and safety.
\item \textbf{Self-optimizing pipelines:} Models that schedule themselves, synthesize kernels, and co-evolve with hardware will appear, reducing design cycles and raising sustained utilization, blurring boundaries between AI model, software stack, and hardware platform. AI-generated and AI-verified kernels will become standard, enabling performance portability and correctness across diverse hardware backends.
\item \textbf{Divergent privacy postures:} Ecosystems are increasingly split between strict on-device inference and secure cloud execution, driving dual-track tooling and deployment strategies. Consumer and regulatory pressure will push edge devices toward stronger local autonomy, while enterprises will consolidate high-value workloads in encrypted, auditable clouds.
\item \textbf{Decentralized and agent-centric AI systems} will increasingly complement centralized cloud-based models. Rather than relying on monolithic inference endpoints, future AI systems may consist of large populations of semi-autonomous agents operating across edge devices, robots, virtual environments, and digital twins, coordinating through sparse communication and shared abstractions. Such decentralized AI ecosystems resemble metaverse-scale systems, where computation, learning, and decision-making are distributed across heterogeneous nodes with varying capabilities and trust assumptions. This paradigm introduces new challenges in orchestration, consistency, security, and energy efficiency, but also offers resilience, scalability, and locality advantages. 
\end{itemize}

\subsection{Potential Obstacles, Pitfalls, and Solutions}

The following issues are identified to be the potential obstacles and pitfalls, and we offer some initial solutions. More are needed from the research community.

\begin{itemize}
\item \textbf{A “chicken-and-egg” problem often stalls progress across fragmented layers of the technology stack (services, systems, hardware):}
\item [] Solutions: Addressing this requires intentional cross-layer collaboration and/or vertical co-investment among service providers, system designers, and SoC and memory vendors.
\item \textbf{Heterogeneity introduces significant software burden and fleet-level trade-offs, further reinforced by brand inertia (“everything is a GPU”):}
\item [] Solutions: Develop common intermediate representations (IRs) and graph compilers, portability layers, and verification toolchains; plan power and capital expenditure (CapEx) allocation strategically; and establish disclosure standards and communication practices that surface real architectural properties, such as dataflow patterns, memory intensity, and interconnect characteristics.
\item \textbf{Legal / IP \& data barriers for distillation:}
\item [] Solutions: Open data trusts; government-backed licensing; provenance/compliance; fund open teacher models explicitly permitting distillation.
\item \textbf{Quantum energy and control efforts are hindered by noise, limited scalability, and integration complexity:}
\item [] Solutions: Invest in cryo-CMOS controls and distributed cryoplants; co-locate HPC / GPU clusters with quantum processors; develop robust control theory and energy-aware hardware–software optimization.
\end{itemize}

\subsection{What Does Success Mean in 10 Years?}

We think that the following goals or milestones can serve as good guidelines to measure the success in 10 years in scalable AI+HW innovation. 

\begin{itemize}
\item \textbf{Systems that reliably execute complex tasks from human intent (HAI realized in practice).}
\item [] Algorithmic success requires reproducible, multi-metric evaluation—covering quality, latency, energy, cost, and utilization—as well as AI systems that can plan effectively, select appropriate tools, models, and resources, enforce safety and verification constraints, and execute multi-step objectives with minimal supervision across cloud, enterprise, and edge or physical AI environments.
\item \textbf{$>$100× end-to-end energy efficiency and $\ge$60\% sustained cluster utilization, optimized at gigawatt scale.}
\item [] Efficiency gains come from compute–memory integration (near / in-memory, 3D-stacked memory, locality-first algorithms) and closed-loop fleet optimization that integrates telemetry, automated tuning, and intelligent scheduling across gigawatt-scale data centers and large edge fleets.
\item \textbf{Fully interoperable heterogeneous systems with seamless orchestration and optical global links.}
\item [] Production stacks integrate CPUs, GPUs, coarse-grain reconfigurable fabrics, domain-specific ASICs, and quantum computers where appropriate; are memory-centric by default; employ dense local and sparse global connectivity with optical networking at the global tier; and orchestrate workloads portably across vendors and sites.
\item \textbf{A mature ecosystem of domain-tuned SLMs distilled from permissively licensed open teachers, deployed across multi-agent frameworks.}
\item [] Lawful, open teacher models with clearly defined usage rights enable domain-specific distillation, while compact small language models (SLMs) operate on edge and robotics platforms to meet privacy, latency, and energy constraints. These SLMs can coordinate with cloud-based LLMs \cite{anthropic2025CLAUDE, google2025GEMINI3, singh2025GPT5} through multi-agent systems that dynamically select the appropriate model, context, and compute resources for each task.
\item \textbf{Self-improving AI-for-chips and systems pipelines delivering $\ge$ 3× faster silicon design cycles with predictable PPA, and mechanistically informed, performance-portable, verified kernels.}
\item [] Models routinely generate and verify kernels, co-design training and inference stacks, and assist across the EDA flow from specification to RTL, verification, closure, and bring-up, with human-in-the-loop sign-off; kernels are performance-portable across the heterogeneous stack.
\end{itemize}

\subsection{Suggested Action Items for Academia, Industry, Government, and the Community}

Finally, we would like to recommend the following actions for academia, industry, and the community for achieving 1000X gain on AI+HW efficiency. 

\begin{itemize}
\item \textbf{Academia:}
\begin{itemize}
\item Focus effort on abstraction-centric learning paradigms (compositional reasoning, program-like intermediates, planning modules, etc.) to move beyond brute force approaches.
\item Advance mechanistic interpretability tied to compression / systems wins.
\item Pursue compute-in / near-memory 3D integrated hardware architecture with small-world interconnectivity topologies for scalable bandwidth.
\item Build objective, task-grounded metrics (quality / latency / energy / memory intensity) and balanced curricula across all paradigms.
\end{itemize}
\item \textbf{Industry:}
\begin{itemize}
\item Co-invest in memory-centric prototypes, distillable foundation teachers, and unified optimization toolchains from telemetry through LLM-generated kernels and verification to rollout.
\item Standardize agent interoperability and resource-selection protocols, such as A2A or MCP (Model Context Protocol), so agents can discover models/resources across domains.
\item Publish utilization / efficiency telemetry; adopt disclosure standards that highlight architectural differences beyond the prevailing “GPU” paradigm.
\end{itemize}
\item \textbf{Government \& Standards Bodies:}
\begin{itemize}
\item Launch DARPA-style SLM challenges; fund open frontier models with explicit distillation rights.
\item Broker access to IP libraries and PDKs (toolbox-style) for legal training data; establish IP / data frameworks (open trusts, provenance).
\item Fund open testbeds for near / in-memory, 3D integration, heterogeneous runtimes, optical interconnects; set energy-efficiency procurement targets; promote architecture-disclosure standards.
\end{itemize}
\item \textbf{Community (All):}
\begin{itemize}
\item Create shared datasets and benchmark kernels for memory- and retrieval-intensive workloads, and systematically track energy consumption per answer.
\item Promote reproducible multi-metric reporting (quality, latency, energy, cost, utilization).
\item Develop agentic evaluation where systems choose models / resources / context under real constraints.
\end{itemize}
\end{itemize}

\section{AI+HW in Action: Applications and Societal Impact}

\subsection{Key Insights}

AI+HW co-design will enable not only faster systems, but fundamentally new classes of applications, ranging from agentic AI and autonomous discovery to real-time interaction with the physical world, that are infeasible under today’s energy and cost constraints. Over the next decade, advances in AI software and hardware will fundamentally transform productivity across nearly every sector. More efficient AI models, coupled with specialized accelerators and memory-centric architectures, will enable real-time reasoning, perception, and control at unprecedented scale and affordability. In industry and the economy, this will translate into smarter design and engineering tools, accelerated innovation cycles, optimized supply chains, resilient manufacturing systems, and autonomous platforms that operate safely and efficiently in real-world environments. In education and workforce development, AI-powered personalized learning systems running on energy-efficient hardware will provide adaptive instruction, continuous reskilling, and accessible education at scale, helping workers transition into new roles as AI-driven automation and physical systems reshape labor markets.

Furthermore, cross-layer AI+HW co-design will be essential for scaling AI into the physical world, where systems must operate under strict real-time, safety, energy, and reliability constraints. Unlike purely digital workloads, physical AI applications tightly couple perception, decision-making, and control with hardware execution, making end-to-end co-design across models, runtimes, and platforms a prerequisite rather than an optimization.

Energy-efficient AI will enable large-scale climate modeling, materials discovery, and optimization of renewable energy and smart grids, as well as real-time monitoring and control of physical infrastructure, without unsustainable carbon footprints. In science and health \cite{Duarte_2018, Edwards_2025,Stokes_2020}, AI accelerators will power faster drug discovery, precision medicine, advanced medical imaging, and continuous health monitoring through wearable and embedded devices. Secure, reliable, and privacy-aware AI systems will also strengthen cybersecurity, critical infrastructure protection, and national security, particularly as autonomous and physical AI systems become more prevalent. By aligning AI software innovation with hardware advances that maximize intelligence per joule, the next decade can unlock transformative societal benefits while ensuring that AI growth remains economically viable, environmentally sustainable, and broadly beneficial.

While “AI and Hardware in Action” might appear to be primarily an industrial concern, academia, industry, and government each have important and complementary roles in building more capable, efficient, and globally competitive AI systems for the United States and the world. In particular, much of industry’s effort is focused on advancing the dominant paradigm of large language models and hyperscale datacenter infrastructure. This strong engineering focus, however, often leaves limited room for exploring fundamentally new directions or rethinking existing assumptions—areas where academia can make distinctive contributions. At the same time, the AI ecosystem spans many companies and multiple layers of the technology stack, and coordination across these actors is limited. Here, government can play a constructive role by encouraging collaboration, aligning priorities, and fostering initiatives that advance shared national and societal interests.

This section on AI and hardware in action highlighted a few issues that are at the heart of facilitating the deployment of AI solutions, the infrastructure crisis, and the incentives for open-ended academic research that is essential for long-term success. In particular, it addresses: (1) the challenge emerging from the gap between piloting a tool and achieving sustained, long-term adoption, complicated by issues of global data sovereignty and strict regulatory compliance, which can potentially slow down innovation, (2) the massive cost and power requirements of large (frontier) AI models, together with the impending power crisis in the US due to insufficient infrastructure and policy action, threaten to slow the adoption of transformative AI technologies, (3) energy efficiency and system scalability are prerequisites for equitable access, preventing AI progress from being limited to a small number of hyperscale actors, (4) Human–AI collaboration will shift human roles toward intent specification, orchestration, and ethical oversight, reshaping engineering practice and workforce training, and (5) bridging the gap between open-ended academic research and incremental industry development, particularly addressing the trade-offs between short-term industrial research vs. academic work that often lacks the necessary scale and focus needed for industry relevance.

\subsection{Key Questions and Answers}

\noindent \textbf{Q1. What is the biggest barrier to real-world AI deployment?}

\noindent The adoption gap remains severe: only about 5\% of piloted AI technologies ultimately translate into sustained financial returns \cite{Challapally_2025}, due to limited continual learning from real-world context, data silos and fragmented data-sovereignty regimes, high operational costs, and regulatory complexity that slows infrastructure development.

\

\noindent \textbf{Q2. What is the most urgent infrastructure challenge?}

\noindent The impending power crisis. Datacenter demand is rising by tens of gigawatts while U.S. generation and grid capacity lag far behind, with China currently holding a substantial power advantage. Without action, we face shortages within 5 years that will constrain AI deployment. We need policies ensuring the 85\% of datacenters operating at <30kW / rack aren't left behind.

\

\noindent \textbf{Q3. How can we bridge the academia-industry gap?}

\noindent Three top-level mechanisms are suggested: (1) collective university–cloud partnerships facilitated and/or negotiated by government agencies; (2) industry-sponsored research programs with sustained funding beyond one-time grants, focused on ambitious long-term initiatives such as new energy-efficient computing paradigms, hardware–software–application co-design, and advanced manufacturing; and (3) academic incentive structures that recognize and reward long-term, practical, and system-level contributions.

\

\noindent \textbf{Q4. How do we prevent growing inequalities in AI hardware access?}

\noindent Preventing growing inequalities in AI hardware access requires efforts across both models and infrastructure. On the model side, developing smaller, efficient, specialized models, such as systems with 20B (or less) active parameters that can run on edge or modest on-premise hardware, can broaden accessibility beyond hyperscale datacenters \cite{Saad-Falcon_2025}. On the hardware side, access must be expanded through open-source tooling and shared infrastructure. Broader access to advanced semiconductor design capabilities, such as EDA tools, fabrication platforms, and advanced PDKs, along with cross-industry benchmarks can further help ensure that AI hardware innovation remains accessible to academia, startups, and emerging research communities.

\

\noindent \textbf{Q5. Is the 1000x efficiency improvement realistic in the next decade?}

\noindent We predict 100x gains in 5 years (high confidence) and 1000x in 6-10 years (moderate confidence) through combined model, software, and hardware advances. \textbf{A viable path toward achieving the ultimate 1000× efficiency improvement is to combine approximately 10× gains from algorithm and model optimization, 20× improvements in silicon utilization and advances, and 5× improvements from system-level efficiency.} These gains can be measured in terms of intelligence per joule. 

\subsection{Important Future Trends}

\noindent\textbf{The Power Crisis Timeline:} Hyperscalers, established tech companies, and startups are commissioning tens of gigawatts of datacenter capacity without matching increases in generation or grid delivery. We predict a U.S. power shortage within 5 years that will constrain AI deployment. Currently, the U.S. substantially trails China in available power, and market forces alone will not solve this crisis (refer to Figure 4).

\begin{figure}
    \centering
    \includegraphics[width=0.8\textwidth]{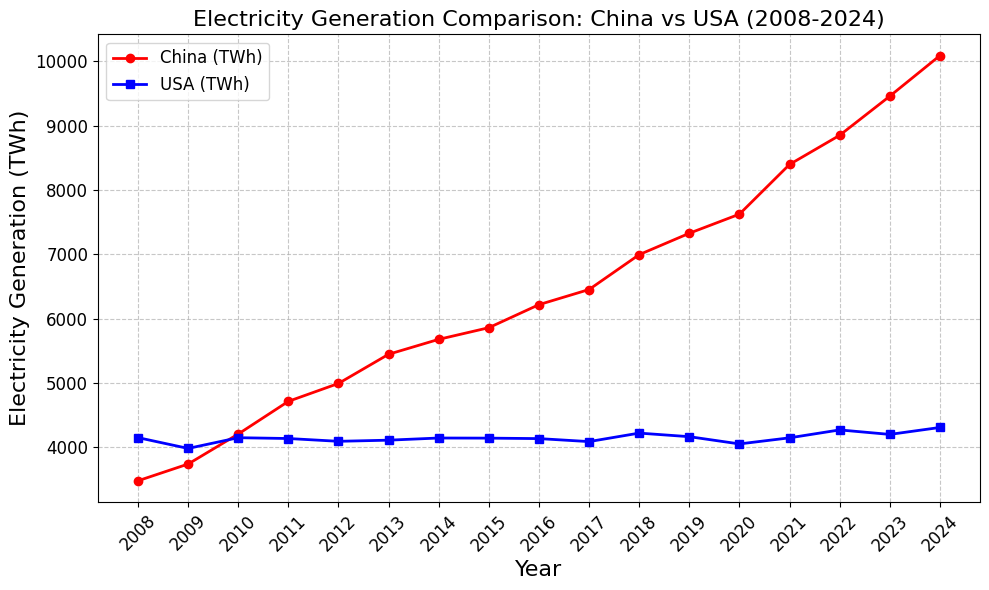}
    \caption{Data Collected from \url{https://en.wikipedia.org/wiki/Electricity_sector_in_China} and \url{https://en.wikipedia.org/wiki/Electricity_sector_of_the_United_States}}
    \label{fig:4}
\end{figure}

\

\noindent\textbf{Cloud-to-Edge Shift:} Currently, cloud is pretty much the only way that the modern AI algorithms are served. We predict a fundamental distribution shift where frontier models give way to smaller, specialized models (<20B parameters) optimized for narrow tasks. These models will migrate from cloud datacenters to edge and end-user applications, such as autonomous vehicles, robotics, and consumer devices, due to their modest resource requirements and improved efficiency. This distribution shift would also increase the number of diverse overall custom chips. 

\

\noindent\textbf{AI-Driven Business Models:} While specific winners remain unpredictable (analogous to the Internet around 1998), we expect multiple successful AI-driven business models within 10 years. Today's recommender systems are established; autonomous vehicles, robotics, agentic AI, and customer service automation show strong promise.

\

\noindent\textbf{Competitive Dynamics:} Cross-sector benchmarks and shared infrastructure will increasingly determine national and global competitiveness in AI deployment. The 1000x efficiency gains will be broadly shared across global technical ecosystems, not exclusive to the US. However, power capacity constraints create competitive risk, with limited U.S. power budgets, a rival state might afford 10x more inference capability even with equally efficient technologies. 

\

\noindent\textbf{Beyond terrestrial infrastructure:} Space-based AI computing represents an emerging and largely unexplored design point for long-term AI infrastructure. Concepts under active consideration by industry envision orbital or near-space platforms where solar energy is abundant and thermal dissipation follows different physical constraints. While such systems are unlikely to replace ground-based datacenters, they open new research opportunities in energy-abundant yet latency-constrained computing, fault-tolerant autonomous operation, radiation-hardened AI hardware, and delay-tolerant learning and inference pipelines. Academic research can play a critical role in defining architectures, algorithms, and control strategies suitable for these regimes, including intermittent execution, extreme autonomy, and physics-aware system optimization. As AI infrastructure planning extends toward 2035, space-based and non-terrestrial computing should be considered as complementary platforms that stress-test our assumptions about energy, reliability, and system design.

\subsection{Potential Obstacles, Pitfalls, and Solutions}

\noindent\textbf{Infrastructure Bottlenecks:} Insufficient U.S. power generation and grid capacity to support the tens to hundreds of gigawatts of new datacenter demand, combined with permitting processes that can take years while global competitors move more quickly.

\noindent\textbf{Solution:} Immediate government investment in alternative energy sources, including small modular reactors (SMRs) with a 5–10 year deployment horizon; streamlined regulatory frameworks for datacenter and energy infrastructure; and proactive public engagement to build support for nuclear and renewable energy deployments.

\

\noindent\textbf{Fragmented Ecosystem:} Insufficient interoperability between hardware, software, and models creates a fractured landscape; siloed disciplinary frameworks in computer engineering, systems, and AI prevent holistic solutions.

\noindent\textbf{Solution:} Cross-industry benchmarks; standard cross-stack performance measurements; multi-stakeholder forums to develop shared infrastructure; government-funded research requiring collaboration across the full stack rather than individual layers.

\

\noindent\textbf{Misaligned Incentives:} Academia is often less incentivized to address practical problems, while industry tends to focus on incremental improvements within existing paradigms.

\noindent\textbf{Solution:} Reshape academic incentives to further value practical engineering and systems-level contributions; industry-sponsored sustained partnerships that align research with real challenges; collective university-cloud agreements to democratize access.

\

\noindent\textbf{Overemphasis on Frontier Models:} Disproportionate focus on AGI-scale frontier models diverts attention from smaller, specialized, edge-centric models that could deliver near-term value and broader access.

\noindent\textbf{Solution:} Research funding specifically targeting efficient small models (models that fit on local hardware); benchmarks emphasizing deployment viability, not just performance on standard tasks; venture capital education on edge-deployment opportunities.

\

\noindent\textbf{Validation Challenges:} New hardware design methodologies cannot be validated without access to realistic libraries, simulation tools, and measurement data; academia lacks industry-scale infrastructure; simulation tools don't match real-world conditions.

\noindent\textbf{Solution:} Industry sharing of anonymized test and measurement data; government-funded "simulation superhighway" validated against real systems; sustained industry partnerships providing access to production environments.

\

\noindent\textbf{Talent and Knowledge Gaps:} U.S. science and engineering talent is insufficient to support current growth; restrictive immigration policies risk losing the international talent that forms the majority of leading tech companies' workforce.

\noindent\textbf{Solution:} Immigration policies that attract and retain top global talent from all over the world, including China, India, and Europe; multi-university collaborations that pool specialized expertise; industry-academia rotations to transfer knowledge bidirectionally.

\

\noindent\textbf{The Risk of Moore's Law Ending:} With Dennard scaling ended, we can mostly only scale out (more chips) rather than up (faster chips), exacerbating power concerns and limiting efficiency gains from traditional approaches.

\noindent\textbf{Solution:} Co-design of algorithms, software, and specialized hardware; hierarchical memory systems that reduce data movement; 3D integration, compute-in-memory or
near-memory, and research into novel computing paradigms beyond CMOS scaling.

\subsection{What Does Success Mean in 10 Years?}

Success means solving the power crisis before it becomes insurmountable, achieving the predicted efficiency gains that enable transformative applications, and maintaining U.S. competitiveness and public trust through coordinated, multi-stakeholder actions. Success also includes open-source tools, shared benchmarks, and accessible infrastructure that allow universities, startups, and smaller institutions to contribute meaningfully to AI innovation. More details below. 

\

\noindent\textbf{Power Crisis Resolved:} Sustainable datacenter scaling through diversified energy infrastructure, including SMRs and alternative sources, with streamlined permitting that enables rapid deployment while maintaining environmental responsibility.

\

\noindent\textbf{Transformative Efficiency Gains:} 1000x improvement in AI efficiency, which in turn would enable a fundamental redistribution of AI traffic from cloud to edge. A large distribution of AI workloads will belong to specialized and smaller models that are deployable on autonomous vehicles, robots, and consumer devices such as AR / VR headsets.

\

\noindent\textbf{Thriving Cross-Sector Ecosystem:} Positive-sum and productive relationships across academia, industry, venture capital, and government. Multi-stakeholder forums drive shared infrastructure (e.g., "simulation superhighways"), advanced benchmarking, open-source tooling, and coordinated research that bridges short-term industrial needs with long-horizon academic innovation.

\

\noindent\textbf{U.S. Competitive Edge:} Despite globally shared efficiency advances, sufficient U.S. power capacity and talent pipeline to match or exceed rivals in AI and hardware capability and innovation velocity.

\

\noindent\textbf{Equitable Access:} Open-source tools, cross-industry benchmarks, and more advances on small, yet capable models such that more players, rather than just massively-funded labs and hyper-scalares can meaningfully and sustainably contribute to the frontier of AI and hardware development.  

\

\noindent\textbf{Flourishing AI-Driven Economy:} A thriving AI-driven economy will be supported by sustained academia–industry–government collaboration rather than isolated, short-term advances. Multiple successful business models and applications will emerge across autonomous vehicles, robotics, agentic AI, and yet-to-be-imagined domains, generating broad economic value comparable to, and ultimately exceeding, the Internet’s post-1998 transformation. 

\

\noindent Ultimately, success will be measured by the ability to deploy intelligent systems at scale in real-world environments—scientific, industrial, and societal—where cross-layer co-design enables AI systems that are efficient, trustworthy, adaptable, and aligned with human and environmental constraints.

\subsection{Suggested Action Items for Academia, Industry, Government, and the Community}

\textbf{Government:} Invest in energy infrastructure, including alternative sources of energy such as small modular reactors (SMRs); streamline datacenter permits; invest in energy and infrastructure research now (5-10 year horizon); and set policies to ensure that existing large-scale data centers not originally designed for AI are not left behind and can be effectively repurposed to support AI workloads. 

It is well established that every \$1 invested in R\&D can return roughly \$5 to the economy \cite{Benjamin_2020}. Strategic investment in large-scale, multi-university research across the chip and system stack, such as open-source EDA tools and generalizable, modular architecture simulators, can significantly amplify this impact. In parallel, facilitate collective negotiations between universities and cloud providers to improve access and efficiency; create opportunities for public engagement and informed discourse; and address the environmental impacts of expanding energy infrastructure while highlighting workforce and job opportunities associated with nuclear power plants and data center development. In addition, establish policies that maintain or strengthen the United States’ leadership in attracting science and engineering talent from around the world. For decades, the U.S. has successfully drawn top talent from regions such as China, India, and Europe to support the workforce of its leading technology companies, and it is critical to preserve this advantage.

\

\noindent \textbf{Universities:} Work with government agencies like NSF or DOE to negotiate cloud partnerships collectively rather than building individual, quickly outdated facilities; focus research on ecosystem-enabling tools like open-source EDA; establish academic incentives for industrial partnerships; focus on long-horizon disruptive algorithmic research over incremental; enable multi-university collaborations on major challenges to pull together resources and interdisciplinary specialization.

\

\noindent \textbf{Industry:} Consider sponsoring and guiding academic works to better align them with industry problems and focus them on solving top challenges. Participate in academic committees to value practical work; provide sustained partnerships beyond one-time grants; validate simulation tools with realistic experiments. Expose test and measurement data and advanced benchmarks \cite{vishwakarma2024QISKITHUMANEVAL} that can enable academia to validate their results. Create cross-industry benchmarks to better drive research and development of AI hardware and software. 

\

\noindent \textbf{Community:} Establish multi-stakeholder forums between academia, industry, venture capital firms, and government to develop shared infrastructure like a "simulation superhighway" for exploring new architectures.

\section{Conclusion and Call to Action}

The next decade will determine whether artificial intelligence evolves from today’s widely adopted digital tools into a foundational technology platform that enables entirely new classes of applications. These include physical AI systems such as robotics, autonomous infrastructure, intelligent manufacturing, and embodied agents, as well as breakthroughs in scientific discovery, healthcare and biomedical research, climate and energy systems, advanced materials design, and large-scale digital infrastructure. Achieving this transformation will require far more than simply scaling model size or deploying denser compute. Instead, AI must advance toward systems that are dramatically more efficient, trustworthy, and deployable across cloud, edge, and real-world environments. Realizing this vision demands deep co-design across algorithms, hardware architectures, and system software. The AI+HW 2035 vision therefore calls for uniting these communities to redefine what scaling means—delivering greater intelligence, adaptability, and real-world impact while reducing energy, cost, and system complexity.

Meeting this challenge requires AI and HW to co-evolve through deep cross-layer collaboration and a fundamental rethinking of system design. By integrating physical, algorithmic, and societal dimensions of intelligence, we can unlock transformative outcomes: 1000× improvements in training and inference efficiency, dramatically improved design productivity, and the establishment of a resilient, sustainable AI infrastructure that advances science and society. This effort must embed human-centric and ethical principles—safety, transparency, accountability, fairness, and societal responsibility—as first-class design constraints rather than afterthoughts. Achieving this is not only a technological objective but also a responsibility to ensure that AI’s growth benefits humanity while minimizing environmental impact.

Meaningful progress will require coordinated and clearly defined action across academia, industry, and government. Academia must develop the foundational theories, abstractions, benchmarks, and open-source platforms that enable rigorous AI+HW co-design, while educating the next generation of researchers fluent across algorithmic and system layers. Industry must translate these advances into scalable, production-ready platforms across cloud, edge, and physical AI systems, investing in deployment, reliability engineering, and real-world validation at scale. Government must catalyze long-term, high-risk research; sustain shared infrastructure such as advanced compute testbeds; align cross-agency priorities; and cultivate inclusive workforce pipelines that broaden participation in AI-enabled innovation. When these sectors act in concert, they can build computing systems that are not only more intelligent and energy-efficient, but also more reliable, secure, and socially responsible.

In essence, AI and HW must co-evolve as an integrated ecosystem, advancing capability, efficiency, and trustworthiness while remaining grounded in human needs and societal context. This is not merely a technical agenda; it is a generational opportunity to redefine how intelligence is built and deployed. Through sustained collaboration, bold investment, and principled innovation, the AI+HW co-design movement can define the next era of computing—one in which intelligence is not only more powerful, but more efficient, reliable, and aligned with the long-term interests of humanity.

To realize this vision, we put forward the following recommended action items.

\begin{itemize}
\item Establish dedicated AI+HW co-design and co-development programs that elevate hardware as a first-class driver of the next AI revolution, rather than treating it as a downstream optimization layer.
\item Launch a national AI+HW initiative (e.g., NSF-led with participation from DARPA, DOE, and NIH) focused on cross-layer research spanning algorithms, architectures, systems, and applications, addressing the full AI stack from models to silicon.
\item Create shared AI+HW infrastructure and resource programs, similar in spirit to NAIRR, that provide academia with access to advanced compute, emerging accelerators, chip prototyping platforms, and system-level testbeds essential for meaningful AI+HW research.
\item Strengthen academia–industry–government collaboration mechanisms, including co-funded research centers, joint fellowships, visiting researcher programs, and shared testbeds, with industry engaged as active stakeholders rather than only external advisors.
\item Issue a call for AI+HW Institutes or Centers, modeled after successful programs such as JUMP, with long-term funding horizons, clear translational goals, and strong workforce-development components.
\item Invest in AI+HW workforce training and education, supporting interdisciplinary curricula, hands-on training with real hardware platforms, and cross-training of AI researchers in systems and hardware, and hardware researchers in modern AI methods.
\item Prioritize system-level evaluation metrics in funded research, including intelligence per joule, intelligence efficiency, data-movement efficiency, real-world utilization, robustness, and deployability—rather than relying solely on model accuracy or peak hardware performance.
\item Address the widening resource-access gap between academia and industry by ensuring that publicly funded researchers have equitable access to large-scale compute, advanced hardware platforms, and realistic datasets.
\item Encourage cross-agency coordination (e.g., NSF, DARPA, NIH, DOE) to align AI+HW investments with national priorities in scientific discovery, healthcare, energy, security, and physical AI systems, including exploring strategic collaboration with emerging DOE initiatives such as the Genesis Mission.
\item Use this report and its arXiv version as a living reference, capturing ongoing community input and guiding future solicitations, policy directions, and coordinated funding programs in AI+HW co-design.
\item Engage professional societies (ACM, IEEE, USENIX, AAAI, ASME, and others) to help advance these actionable items through community building, standards, and advocacy.
\item Engage policymakers and legislators, using this report to inform and educate on the strategic importance of AI+HW co-design for national competitiveness and societal impact.
\item Engage leading industrial partners, sharing this vision to form strategic alliances and pursue joint initiatives aligned with mutual interests.
\end{itemize}

\section{Acknowledgment}

This paper emerged from an NSF-sponsored workshop with the same title, during which each co-author presented insights and perspectives on AI+HW co-design aimed at shaping the combined landscape of these two closely intertwined domains over the next decade. We acknowledge NSF and the following NSF program directors---Sharon Hu, Sankar Basu, Andrey Kanaev, and Raj Acharya---for their support of this workshop. We also thank Mr. Jinghua Wang of the University of Illinois Urbana--Champaign for his assistance in creating several figures and in polishing the writing and formatting of this paper. The presentation slides and summaries of the breakout discussion sessions at the workshop are available at the following website: \url{https://publish.illinois.edu/ai-hw-workshop/schedule/}.

\printbibliography

\end{document}